\documentclass[10pt,journal,compsoc]{IEEEtran}
\usepackage{graphicx}
\usepackage{booktabs} 
\usepackage{caption}
\usepackage{subcaption}
\usepackage[titlenumbered, ruled,vlined]{algorithm2e}
\usepackage{tikz}
\usepackage{amsmath}
\usepackage[adobe-utopia]{mathdesign}
\usepackage{array,multirow,graphicx}
 \usepackage{float}
 \usepackage{amsfonts}
 \usepackage{color}
\usepackage{subcaption}
\usepackage[noend]{algpseudocode}
\usepackage{color, colortbl}
\usepackage{xspace, xcolor}
\usepackage[capitalize]{cleveref}
\usepackage{mwe}
\usepackage{pgfplots}
\usepackage{enumitem}
\usepackage{todonotes}

\ifCLASSOPTIONcompsoc
  \usepackage[nocompress]{cite}
\else
  \usepackage{cite}
\fi
\ifCLASSINFOpdf
\else
\fi
\newcommand*{\Comb}[2]{{}^{#1}C_{#2}}
\begin{document}
%
\title{Adaptive Neural Message Passing for Inductive  Learning on Hypergraphs}
%
%
%
%

\author{Devanshu Arya,
        Deepak K. Gupta,
        Stevan Rudinac,
        and~Marcel Worring,~\IEEEmembership{Senior Member,~IEEE}
\IEEEcompsocitemizethanks{\IEEEcompsocthanksitem Devanshu Arya, Stevan Rudinac and Marcel Worring are with the University of Amsterdam, The Netherlands. Deepak Gupta is with Indian Institute of Technology Dhanbad, India and at AIQ, Abu Dhabi, UAE. \protect
E-mail: d.arya@uva.nl
}
}

\IEEEtitleabstractindextext{%
\begin{abstract}

Graphs are the most ubiquitous data structures for representing relational datasets and performing inferences in them. 
They model, however, only pairwise relations between nodes and are not designed for encoding the higher-order relations. 
This drawback is mitigated by hypergraphs, in which an edge can connect an arbitrary number of nodes.  
Most hypergraph learning approaches convert the hypergraph structure to that of a graph and then deploy  existing geometric deep learning methods. 
This transformation leads to information loss, and sub-optimal exploitation of the hypergraph's expressive power. 
We present HyperMSG, a novel hypergraph learning framework that uses a modular two-level neural message passing strategy to accurately and efficiently propagate information within each hyperedge and across the hyperedges. 
HyperMSG adapts to the data and task by learning an attention weight associated with 
each node's degree centrality. Such a mechanism quantifies both local and global importance of a node, capturing the structural properties of a hypergraph. 
HyperMSG is inductive, allowing inference on previously unseen nodes. Further, it is robust and outperforms state-of-the-art hypergraph learning methods on a wide range of tasks and datasets. 
Finally, we demonstrate the effectiveness of HyperMSG in learning multimodal relations 
through detailed experimentation on a challenging multimedia dataset.

\end{abstract}

\begin{IEEEkeywords}
Hypergraphs, Representation Learning, Multimodal Learning, Geometric Deep Learning
\end{IEEEkeywords}}

\maketitle

\IEEEdisplaynontitleabstractindextext

%
\IEEEpeerreviewmaketitle

\ifCLASSOPTIONcompsoc
\IEEEraisesectionheading{\section{Introduction}\label{sec:introduction}}
\else
\section{Introduction}
\label{sec:introduction}
\fi

\begin{figure*}
    \centering
    \includegraphics[scale=0.5]{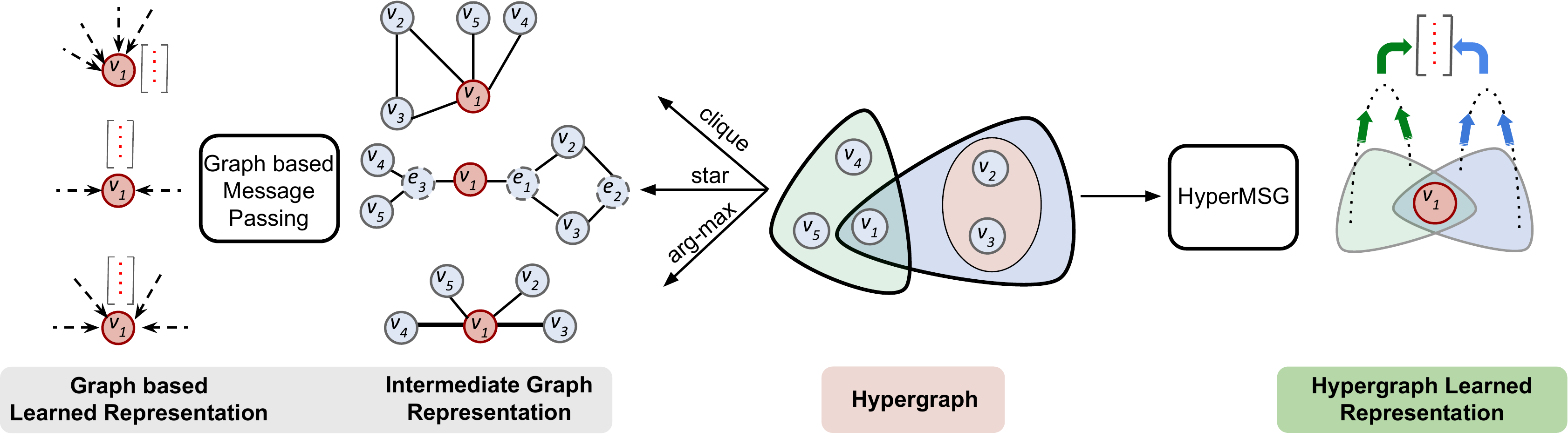}
    \caption{Illustration of the difference between traditional methods that require an intermediate graph representation and our method (HyperMSG). The left side shows the reduction of a hypergraph to a graph using clique, star and functional metric based (arg-max)
    \cite{feng2019hypergraph}\cite{yadati2019hypergcn} expansion methods. The clique expansion loses the unique information associated with the hyperedge defined by the set of nodes $\{v_2, v_3\}$, and it cannot distinguish it from the hyperedge defined by the nodes $\{v_1, v_2, v_3\}$. Star expansion  creates a heterogeneous graph that is difficult to handle using most well-studied graph methods \cite{yang2020hypergraph}. Functional metric (in particular arg-max) based expansion \cite{yadati2019hypergcn} does not exploit the full structural information within hypergraph. On the other hand, HyperMSG formulates neural message passing framework on hypergraphs without any reduction to graphs and directly learns node representations. Hence, it provides a much better basis for hypergraph representation learning. 
}
    \label{fig1}
\end{figure*}

Modelling the intrinsic properties of many real-world datasets, such as their structure and connections between the data points,
requires relational data structures. 
Graphs are a popular data structure for discovering useful information in relational datasets due to their capability to combine object-level information with the underlying inter-object relations \cite{wu2020comprehensive}.
Data  encountered in many real-world scenarios, however, contain relationships among objects which are not dyadic (pairwise) but  triadic, tetradic or higher. 
As an example, consider a research community, where authors publish papers in groups of more than two. Representing such groups of collaborators using just pairwise edges specific to ordinary graphs inevitably leads to  information loss.
In such cases, the interactions among the objects can be fully modelled by including higher-order relations instead of pairwise relations only \cite{wolf2016advantages}. 
Some example domains where graphs are already shown to be insufficiently expressive are social networks \cite{tan2011using}, video surveillance \cite{yan2020learning} and neuroscience \cite{gu2017functional}. In all such domains the information at group level contributes to understanding the data and solving tasks. 
Group level and other higher-order relationships can be readily represented with \emph{hypergraphs} \cite{berge1976graphs}, where a \emph{hyperedge} can connect an arbitrary number of nodes as opposed to just two nodes in a graph. 

Hypergraphs have already been shown to provide a flexible and natural framework to represent higher-order relations within homogeneous data \cite{zhou2007learning}. Real-world datasets, however, often consist of objects with multiple modalities.
Traditional graph- or hypergraph-based representations for multimodal data model relations in each modality independently, ignoring  the heterogeneous relations between objects. Yet, such relations in the multimodal data can provide complementary information revealing fundamental characteristics of objects and their context.
For instance, consider a social media platform which has users, tweets, hashtags, and images. Representing multimodal relations in the collective information generated when a user posts a tweet with an image containing multiple hashtags is infeasible using binary pairwise relations \cite{li2013link} or by considering tweets, users, and images as non-related channels.
Hypergraphs efficiently represent  multimodal objects with higher-order relations inside a channel and among channels in various domains such as  visual arts \cite{arya2019hyperlearn}, discussion forums \cite{arya2019predicting}, visual question answering systems \cite{kim2020hypergraph}, music recommender systems \cite{bu2010music} and protein-protein interaction networks \cite{klimm2021hypergraphs}. 

Apart from being multimodal, in real-world datasets the objects and relations among them are dynamically evolving. That is, during training time there will be nodes which are unseen and relations which are partially observed. Such instances often exist in scenarios such as finding the most promising target audience for a marketing campaign or making movie recommendations with new movies appearing all the time. Performing inference on  unseen nodes is challenging and requires an inductive learning framework.

Making inferences in relational datasets represented on a graph or hypergraph requires means to accurately propagate information across the nodes. 
Recent advances in geometric deep learning \cite{bronstein2017geometric} resulted in several techniques to do so by using spectral or spatial convolution on graphs. Spectral graph convolution is based on spectral graph theory \cite{bruna2013spectral} and provides a well-founded mathematical framework for designing translation-invariant operators (filters). It requires, however, computing eigenvectors of large matrices, which is computationally expensive.
Another shortcoming is that these filters are not localized i.e. to compute the output of a node, the convolution operation does not consider a limited number of neighborhood hops. Hence, its complexity grows with the size of the graph and therefore the method is not scalable to large datasets. This led to works applying approximation to the spectral graph convolution using $K$-localized filters which use a $K$th-order polynomial, like in \emph{ChebNet} \cite{defferrard2016convolutional} and \emph{GCN} \cite{kipf2016semi}. Such filter localization techniques can be interpreted as spatial methods that perform message passing on local neighborhoods which are a certain number of steps away from a node.
Due to their efficiency, spatial graph convolution approaches have received most attention in recent years. 
The resulting neural network model performs message passing operations informed by the structure of the graph to distribute encoded feature information among connected nodes \cite{gilmer2017neural}\cite{kipf2020deep}. Message passing has proven to be effective for graph inference, but has yet to be researched for hypergraphs.

Several approaches for devising message passing in hypergraphs have been proposed aiming to exploit their expressive power \cite{yadati2019hypergcn,feng2019hypergraph,yadati2020neural}. 
A common implicit assumption they make is that a hypergraph is a specific type of ordinary graph. If the assumption holds, reducing the hypergraph learning problem to that of a graph should suffice. 
Strategies to reduce a hypergraph to a graph include transforming the hyperedges into multiple edges using clique expansion \cite{feng2019hypergraph}\cite{jiang2019dynamic}\cite{ zhang2018beyond}, converting  to a heterogeneous graph using star expansion \cite{agarwal2006higher}, and replacing every hyperedge with a weighted edge created using a certain predefined metric on the functional properties of the node  \cite{yadati2019hypergcn}.  
By using a graph as an intermediate (proxy) representation, these approaches make existing graph-based message passing methods \cite{gilmer2017neural,wu2020comprehensive} applicable to hypergraphs, as can be seen in the left part of Fig.~\ref{fig1}.  However, a hypergraph is not a special case of graph. The opposite is true, graphs are simply a specific type of hypergraph \cite{berge1976graphs}, where the hyperedges are a superset of the pairwise edges. The complex relations in a hypergraph cannot be viewed as an instantiation of an ordinary graph, thus reducing the hypergraph problem to that of a graph cannot fully utilize the available information \cite{dong2020hnhn}. To address tasks in datasets with higher-order relations, a truly hypergraph-based message passing formulation is needed that ensures information propagation within and across hyperedges. 

 Another major limitation of the existing hypergraph representation learning frameworks is their inherently transductive nature  \cite{yadati2019hypergcn}\cite{bai2020hypergraph}, thus they are inapplicable for making any inferences on unseen nodes.  
 Often, the injection of new unseen nodes can lead to the introduction of noisy features in the message passing framework \cite{fox2019robust}. 
 We introduce a probabilistic neighborhood sampling approach which acts as a regularizer and facilitates the proposed inductive learning  framework. In a hypergraph, message passing operation on a node is different from regular graphs. This is because in a hypergraph the message propagation from the neighborhood of a node is performed at two levels - within a hyperedge and across hyperedges. Thus unlike regular non-weighted graphs, in a hypergraph not all neighborhood nodes hold equal importance. To quantify the role of a node, we identify that the uniqueness of a hypergraph lies in the detailed structure of its hyperedges as well as the distribution of nodes across them. In a graph, each edge is only shared between two nodes. A hyperedge, however, can contain a large set of nodes, and each node could have different contributions within the hyperedge based on its relations with neighboring nodes. This is often defined by empirical measures, such as the degree centrality \cite{kapoor2013weighted}\cite{benson2019three}. Degree centrality captures how popular or active a node is in a hypergraph. Since such a paradigm is sensitive to the choice of dataset as well as the task, choosing an empirical measure is not optimal. We introduce a deterministic neighborhood attention mechanism which quantifies the importance of a node in terms of its neighborhood and the number of hyperedges to which it belongs. The proposed attention mechanism is self-adaptive to the choice of task, dataset and to any variation in the hypergraph structure. Combining such an attention mechanism with our message passing framework, we propose an inductive learning framework that exploits the full structure of hypergraphs, without any hypergraph-to-graph conversion, to perform several tasks such as node classification, link prediction, and hypergraph classification.

The points below highlight the contributions of this paper.
\begin{itemize}
    \item We present HyperMSG, a hypergraph learning framework with a two-level message passing scheme that
    jointly captures the relations within a hyperedge and across the hyperedges.
    \item HyperMSG is inductive in nature, and facilitates probabilistic sampling of both seen and unseen nodes, based on their importance in message passing.
    \item HyperMSG adapts to the dataset and task by implicitly learning the importance of neighborhood nodes in representation learning.
    \item We demonstrate that HyperMSG outperforms the state-of-the-art methods by remarkable margins on standard benchmarks consisting of citation networks. In addition, we show that HyperMSG is highly efficient in exploring the complex heterogeneous interactions in multimodal hypergraphs to perform tasks such as multi-label classification, link prediction and recommendation. The robustness and general applicability of HyperMSG is further validated on the task of hypergraph (brain) classification in an extremely noisy neuroimaging dataset. 
\end{itemize}

\section{Related Work}
\label{sec:related_work}

Message passing on hypergraphs aims at learning low-dimensional representations for signals (features) defined on the nodes. Recently, several methods have been proposed for message passing on graphs, and deployed for modeling physical systems, learning molecular fingerprints, predicting protein interfaces, and classifying diseases \cite{zhou2020graph}. However, there is a lack of an accurate message passing framework for hypergraphs.  The biggest challenge in hypergraph-based learning is posed by the high variation in hyperedge cardinality i.e. the number of nodes in each hyperedge, which limits the accurate and efficient information propagation from one node to another along the hyperedges. For better understanding of message passing neural networks, we first provide a brief overview of some popular graph-based message passing neural networks.  We will then discuss the existing techniques for representation learning on hypergraphs that are related to our proposed framework HyperMSG.


\textbf{Message Passing in Graphs}  The general idea behind a graph convolutional neural network is to define convolution on graphs, where the input is a graph instead of e.g. a 2-D image grid. 
Methods for performing convolution on graphs can be broadly classified into spatial (message passing) or spectral methods \cite{bronstein2017geometric}.
As mentioned in the introduction, the major drawbacks of spectral approaches is their rigidity in generalizing to new graph structures and their high computational complexity \cite{zhou2018graph,zhang2020deep}. In this work, we focus on spatial approaches that use message passing neural networks. The basic framework of any message passing network takes in a graph with signals (i.e. features) on each of its nodes 
as input and learns embeddings for each node by aggregating information from its neighbors \cite{xu2018powerful}. Message passing neural networks outline a general message passing framework for learning such an aggregation mechanism on graphs. It passes information (messages) from one node to another along edges and repeats it in $K$-steps to let information propagate through the graph. Several variants of this general approach have been proposed, such as \cite{gori2005new,li2015gated, kipf2016semi, gilmer2017neural,hamilton2017inductive}. Recently, it has been further extended to heterogeneous graphs in which nodes (and edges) are typed, allowing graphs to incorporate auxiliary information. Some of these ideas include using attention-based mechanism such as typed attention \cite{linmei2019heterogeneous}, neighbour attention \cite{zhang2019heterogeneous}, vertex-level and semantic-level attention \cite{wang2019heterogeneous} across different types of nodes or using metapath based aggregations \cite{fu2020magnn}. Unlike graph-based models with only binary relations, hypergraph learning models need to explore the higher-order relations in the data \cite{yoon2020much}.

\textbf{Message Passing in Hypergraphs}
Since the introduction of learning with hypergraphs \cite{zhou2007learning}, several such methods have been introduced \cite{gao2020hypergraph}, and successfully deployed in various tasks, such as  link prediction \cite{li2013link}, community detection \cite{chien2018community} and visual object tracking \cite{wen2019learning}. In spectral theory of hypergraphs, methods  have been proposed that fully exploit the hypergraph structure using non-linear Laplacian operators \cite{chan2018spectral,hein2013total}. However, these methods have similar drawbacks to spectral graph methods with regard to their computational complexity and scalability. 
Learning on the hypergraph can also be seen as the process of message passing along the hypergraph structure in analyzing the structured data. Emulating a graph-based message passing framework for hypergraphs is not straightforward since a hyperedge involves more than two nodes which makes the interactions inside each hyperedge complex. Hypergraphs are mathematically represented using either an incidence matrix or an adjacency tensor. Incidence matrix based representations of hypergraphs are rigid in describing the structures of higher order relations \cite{li2013z}. On the other hand, formulating message passing on a higher-dimensional representation of a hypergraph using adjacency tensors makes it computationally expensive 
and restricted to small datasets and uniform hypergraphs \cite{zhang2019introducing}. 

To circumvent the above issues,  \cite{feng2019hypergraph} and \cite{bai2020hypergraph} reduce a hypergraph to a graph using clique expansion and perform graph convolutions on them. Further, \cite{jiang2019dynamic} assume the initial hypergraph structure is weak, and  extend the work of \cite{feng2019hypergraph} to construct dynamic hypergraphs. Recently proposed HyperGCN replaces a hyperedge with pair-wise  weighted edges between vertices called mediators \cite{yadati2019hypergcn}. The weights are calculated by comparing the functional properties of neighboring nodes (\emph{e.g.}, using \texttt{arg-max} over them) and hence lose the structural information within a hypergraph. With the use of mediators, HyperGCN can be interpreted as an improved variant  of clique expansion, and to the best of our knowledge, is also the state-of-the-art method for all the hypergraph representation learning methods, where a graph based message passing neural network is eventually used. However, it still suffers from the same limitation as the clique expansion \cite{dong2020hnhn}. These limitations are further discussed on the example of a Fano plane in the Appendix. 
 Furthermore, these approaches are inherently transductive and thus, as indicated earlier, cannot perform inference on unseen nodes. In \cite{payne2019deep}  a generalized hypergraph learning framework is presented that uses random walks, however its message passing framework is not robust for heterogeneous hypergraphs and its validity for inductive setting is still unexplored.
 None of the above approaches utilize the complete structural information in the hypergraph, leading to sub-optimal learning performance.


In our preliminary work we introduced \emph{HyperSAGE} \cite{arya2020hypersage}, an inductive representation learning framework for hypergraphs that can exploit their full  structure by aggregating messages in a two-stage procedure. 
It managed to achieve near state-of-the art performance on benchmark datasets.  However, it suffers from lack of adaptability to datasets and structural variation in hypergraphs and poor parallelism. Huang et al. \cite{huang2021unignn} further built over our proposed HyperSAGE framework to circumvent these issues and generalize the aggregation approach. Recently, \cite{ding2020more} proposed message passing in hypergraphs using a dual attention mechanism for classifying text in documents by constructing multiple types of hyperedges (sequential, syntactic and semantic). The proposed message passing framework combines both types of neighborhood information (within hyperedges and across hyperedges)  separately and thus mitigates the shortfall of hypergraph to graph conversion. However, the proposed dual attention mechanism uses the functional properties (i.e. node features) of the hypergraph. This leads to insufficient information propagation as it fails to quantify the importance of a node with respect to its neighbors within and across hyperedges, which is an important aspect for capturing the structural properties of a hypergraph. In this work, we propose HyperMSG, which eliminates matrix- or tensor-based formulations in its neural message passing scheme for hypergraphs. It is further inductive and utilizes all the available information in a hypergraph.

\section{Hypergraph preliminaries}
\label{sec:preliminary}
In this section, we first introduce some of the preliminary definitions and notations of hypergraphs which will be used to formulate the HyperMSG framework.

\textbf{Definition 1} (Hypergraph). \textit{A  hypergraph $\mathcal{H}$ is represented as $\mathcal{H}=(\mathcal{V, E, \mathbf{X}})$, where $\mathcal{V}=\{v_1, v_2, ... , v_N\}$ denotes a set of $N$ nodes and $\mathcal{E}=\{\mathit{e}_1, \mathit{e}_2, ... , \mathit{e}_K\}$ a set of hyperedges, with each hyperedge comprising a non-empty subset from $\mathcal{V}$.  $\mathbf{X} \in \mathbb{R}^{N \times d}$ denotes the feature matrix, with the feature vector $\mathbf{x}_i $ corresponding to  the respective $v_i$ column.
}

The cardinality of any hyperedge $\mathit{e}_l$ is the number of nodes contained in that hyperedge, given by $|\mathit{e}_l|$. Unlike in a graph, the hyperedges of $\mathcal{H}$ can contain different number of nodes i.e. $1 \leq |e_i| \leq |\mathcal{V}|$.
Definition 1 makes it clear that graphs are simply a special case of hypergraphs with a fixed cardinality of 2 for all the edges.


We define three different types of neighborhoods as follows. 

\textbf{Definition 2} (Intra-edge neighborhood). \textit{The intra-edge or local neighborhood of a node ${v_i} \in \mathcal{V}$ for any hyperedge $\mathit{e} \in \mathcal{E}$ is defined as the set of nodes $v_j$ belonging to $\mathit{e}$ and is denoted by  $\mathcal{N}(v_i, \mathit{e})$}.

The intra-edge neighborhood of a node captures the higher order relationships and provides localized group level information to it. Further, let $E(v_i) = \{ \mathit{e} \in \mathcal{E}  |  v_i \in \mathit{e}\}$ be the set of hyperedges containing node $v_i$. The degree of node $v_i$ is thus given by $|E(v_i)|$.

\textbf{Definition 3} (Inter-edge  neighborhood). \textit{The inter-edge or global neighborhood of a node ${v_i} \in \mathcal{V}$, is defined as the neighborhood of ${v_i}$ spanning across the set of hyperedges $E(v_i)$ and is given by $\mathcal{N}(v_i) = \bigcup_{\mathit{e} \in E(v_i)}\mathcal{N}(v_i, \mathit{e})$.}

The inter-edge or global neighborhood of a node captures its global positioning and gathers information from hyperedges similar to the node. Finally, the condensed neighborhood parameterized by $\alpha$ gives a subset of nodes within a hyperedge.

\textbf{Definition 4} (Condensed neighborhood).\textit{ The condensed neighborhood of any node ${v_i} \in \mathcal{V}$ is defined as the sampled set $\mathcal{N}{(v_i,\mathit{e};\alpha)}$ comprising of $\alpha$ nodes from a hyperedge $\mathit{e} \in E(v_i)$, if $\alpha < |e|$, or  all nodes in $\mathit{e}$ if $\alpha >= |e|$.}



\section{Proposed Model}
\label{sec:proposed model}
The main concept behind devising a message passing neural network on hypergraphs is to aggregate feature information from the neighborhood of a node which spans across multiple hyperedges with varying cardinality. In this section we propose HyperMSG, a framework that performs message passing at two levels for a hypergraph. Further, we discuss our adaptive framework that implicitly learns the importance of each node in the representation learning process. Our approach inherently allows inductive learning, which makes it also applicable on hypergraphs with unseen nodes. 
\subsection{Two-level Message Passing Framework}
\label{sec_msgpass}
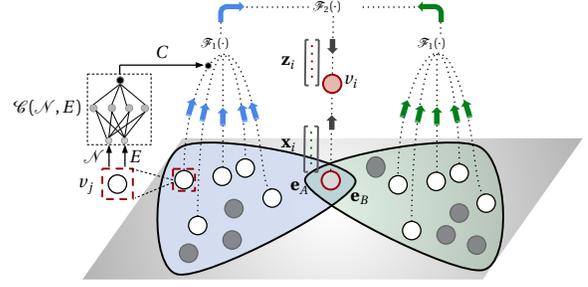
\begin{figure}{}{}
\centering
\begin{tikzpicture}
    \node () at (0,0) {\includegraphics[scale = 0.32]{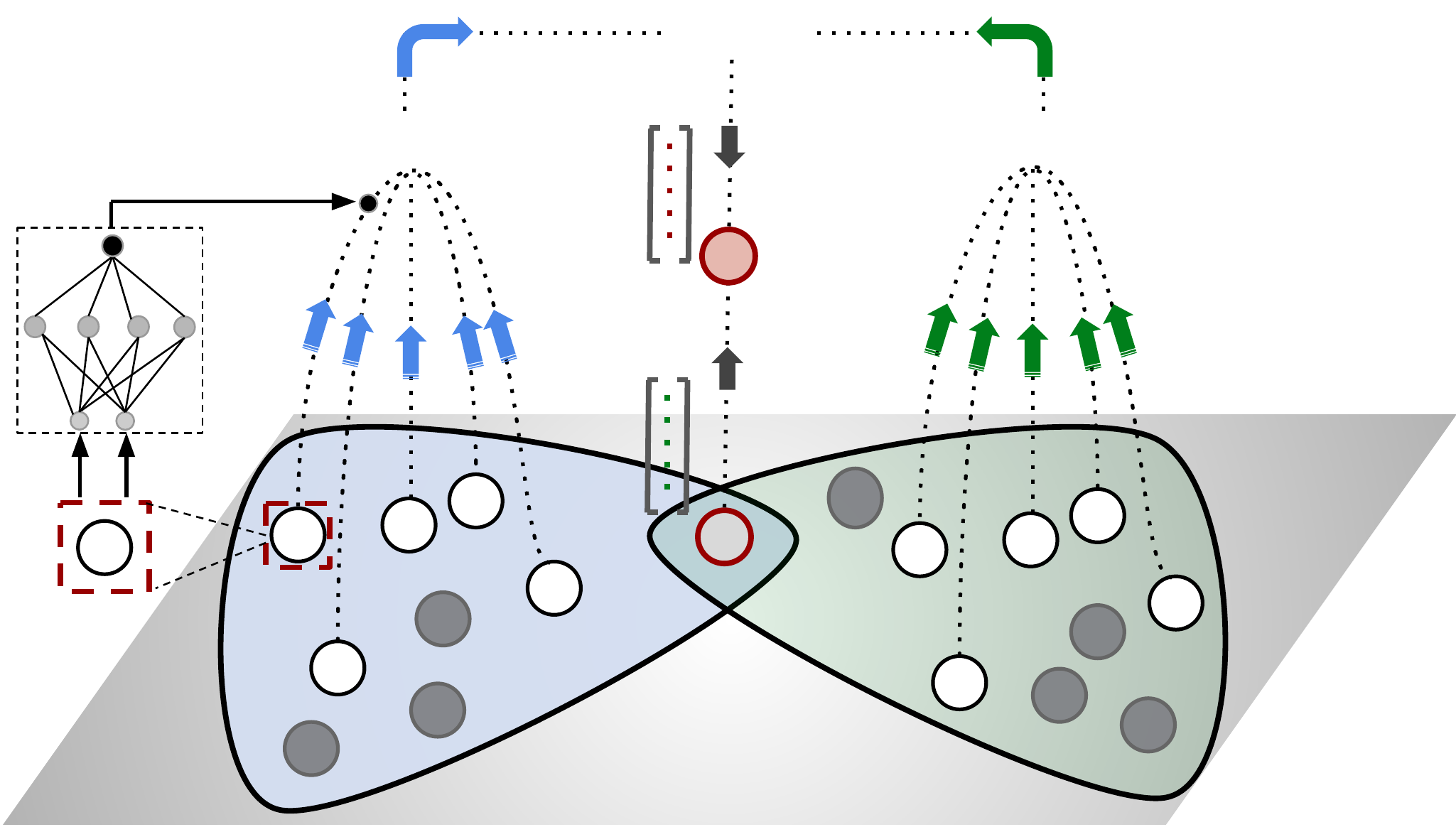}};
    \node[text width=3pt] (whitehead) at (1.19,1.26)
    {{\tiny $\mathcal{F}_1(\cdot)$}};
    \node[text width=3pt] (whitehead) at (-1.69,1.26)
    {{\tiny $\mathcal{F}_1(\cdot)$}};
    \node[text width=3pt] (whitehead) at (-0.20,1.75)
    {{\tiny $\mathcal{F}_2(\cdot)$}};
    \node[text width=3pt] (whitehead) at (-0.62,-0.10)
    {{\scriptsize $\mathbf{x}_i$}};
    \node[text width=3pt] (whitehead) at (-0.62,1.0)
    {{\scriptsize $\mathbf{z}_i$}};
    \node[text width=3pt] (whitehead) at (0.2,0.75)
    {{\scriptsize $v_i$}};
    \node[text width=3pt] (whitehead) at (-0.5,-0.66)
    {{\scriptsize $\mathbf{e}_A$}};
    \node[text width=3pt] (whitehead) at (0.3,-0.8)
    {{\scriptsize $\mathbf{e}_B$}};
    \node[text width=3pt] (whitehead) at (-3.25,-0.24)
    {{\scriptsize $\mathcal{N}$}};
    \node[text width=3pt] (whitehead) at (-2.65,-0.24)
    {{\scriptsize $E$}};  
    \node[text width=3pt] (whitehead) at (-4.2,0.4)
    {{\scriptsize $\mathcal{C}(\mathcal{N},E)$}}; 
    \node[text width=3pt] (whitehead) at (-3.33,-0.64)
    {{\scriptsize $v_j$}}; 
    \node[text width=3pt] (whitehead) at (-2.3, 1.15)
    {{\scriptsize ${C}$}}; 
\end{tikzpicture}
\caption{Schematic representation of the two-level message passing scheme of HyperMSG, with aggregation functions $\mathcal{F}_1(\cdot)$ and $\mathcal{F}_2(\cdot)$. It shows information aggregation from two hyperedges $\mathbf{e}_A$ and $\mathbf{e}_B$, where the intra-edge aggregation is from sampled sets of 5 nodes ($\alpha = 5$) for each hyperedge. The function $\mathcal{C}(\mathcal{N},E)$ is used to compute attention weight $C$ for quantifying the importance of node $v_j$ during intra-edge aggregation on $\mathbf{e}_A$.  For node $v_i$, $\mathbf{x}_i$ and $\mathbf{z}_i$ denote the input and aggregated feature vector, respectively.
}
\label{schematic}
\end{figure}

We propose to interpret the propagation of information in a given hypergraph as a two-level aggregation problem, where the neighborhood of any node is divided into its \emph{intra-edge} neighbors and \emph{inter-edge} neighbors. This information is present in the form of signals on each node often referred to as message as mentioned in the previous section. For message aggregation, we define the aggregation function as a permutation invariant set function on a hypergraph $\mathcal{H}=(\mathcal{V}, \mathcal{E}, \mathbf{X})$ that takes as input a countable unordered message set and outputs a reduced or aggregated message of the same dimension as the original message. Further, for two-level aggregation, let $\mathcal{F}_1(\cdot)$ and $\mathcal{F}_2(\cdot)$ denote the intra-edge and inter-edge aggregation functions, respectively. Schematic representation of the two functions is provided in Fig.~\ref{schematic}. Similar to $\mathbf{X}$, we also define $\mathbf{Z} \in \mathbb{R}^{N \times l}$ as the aggregated feature matrix 
built using the outputs $\mathbf{z}_i$ with dimension $l$ from the aggregation functions. Message passing at node $v_i$ can then be expressed as
\begin{align}\label{eq:3}
    & \mathbf{s}_{1} \leftarrow \{\mathbf{x}_{j}\: | \: v_j \in \mathcal{N}(v_i,\textit{e} ; \alpha)\}, \\
    & \mathbf{s}_2 \leftarrow \{ \mathcal{F}_1^{(\textit{e})}\,(\mathbf{s}_1) \enskip | \enskip \textit{e} \in E(v_i)\}, \\
    & \mathbf{z}_{i} \leftarrow \mathbf{x}_{i} + \mathcal{F}_2(\mathbf{s}_2),
\end{align}
where $\mathbf{s}_1$ and $\mathbf{s}_2$ denote the unordered sets of feature vectors and intra-edge aggregations, respectively. 

To ensure that the expressive power of a hypergraph is preserved or at least the information loss is minimized, the choice of aggregation function should comply with certain properties. First, it should capture the features of neighborhood nodes in a manner that is invariant to permutation of nodes and hyperedges within its neighborhood. This ensures the final message propagation to a node does not depend on the sequence in which we pick up  its neighbors. 
Many graph-based methods use aggregation functions, such as \texttt{mean} and \texttt{max} functions \cite{kipf2016semi}. Mean and max-pooling aggregators are well-defined multiset functions because they are permutation invariant and these functions learn different attributes of the neighborhood (\texttt{max} learns distinct elements and \texttt{mean} learns distributions).  These aggregations have proven to be successful for node classification problems \cite{kipf2016semi, hamilton2017inductive}. However, they are not injective and hence have their limitations in learning unique representations in various cases such as when the nodes have repeating features \cite{xu2018powerful} or when the
features are continuous \cite{corso2020principal}.
There are other non-standard neighbor aggregation schemes that we do not cover, e.g., weighted average via attention \cite{velivckovic2017graph}, LSTM pooling \cite{hamilton2017inductive, murphy2018janossy} and stochastic aggregations \cite{wang2021stochastic}. We emphasize that our theoretical framework is general enough to characterize the representational power using a family of generalized mean aggregation functions. In the future, it would be interesting to apply our framework to analyze and understand other aggregation schemes.


 \textbf{Property 1}  (Hypergraph Isomorphic Equivariance). \emph{A message aggregation function $\mathcal{F}(\cdot)$ is equivariant to hypergraph isomorphism, if for two isomorphic hypergraphs $\mathcal{H} = (\mathcal{V}, \mathcal{E}, \mathbf{X})$,  \mbox{$\mathcal{H}^* = (\mathcal{V}^*, \mathcal{E}^*, \mathbf{X}^*)$}, and permulation operator $\sigma$, given that \mbox{$\mathcal{H}^* =\sigma \bullet \mathcal{H}$}, and $\mathbf{Z}$ and $\mathbf{Z}^*$ represent the aggregated feature matrices obtained using $\mathcal{F}(\cdot)$ on $\mathcal{H}$ and $\mathcal{H}^*$ respectively, 
 the condition \mbox{$\mathbf{Z}^* =\sigma \bullet \mathbf{Z}$} holds.}

Secondly, the aggregation function should also preserve the global neighborhood invariance at the `dominant nodes' of the graph. Here, dominant nodes refer to nodes that act as hubs in power-law networks, possessing many more connections than their neighbors \cite{albert2002statistical}.
The aggregation function should ideally be insensitive to whether the provided hypergraph contains a few large hyperedges, or a large number of smaller ones obtained from splitting them. Generally, a hyperedge would be split in a manner that the dominant nodes are shared across the resulting hyperedges. In such cases, global neighborhood invariance would imply that the aggregated output at these nodes before and after the splitting of any associated hyperedge stays the same. Otherwise, the learned representation of a node will change significantly with each split.
Based on these considerations, we define the following properties for a generic message aggregation function. 

\textbf{Property 2} (Global Neighborhood Invariance).
\emph{A message aggregation function $\mathcal{F}(\cdot)$ satisfies global neighborhood invariance at any node $v_i \in \mathcal{V}$ for a given hypergraph $\mathcal{H} = (\mathcal{V}, \mathcal{E}, \mathbf{X})$, if for any hyperedge $\mathbf{e} \in E(v_i)$ being split into multiple hyperedges without changing $\mathcal{N}(v_i)$, and $\mathbf{z}_i$ and $\mathbf{z}_i^*$
denoting the aggregated feature vectors obtained before and after splitting, the condition  $\mathbf{z}_i^* = \mathbf{z}_i$ holds. }

\textbf{Aggregation using Generalized Means. }One major advantage of our strategy is that the message passing module is decoupled from the choice of the aggregation method itself. This allows our approach to be used with a broad set of generalized means aggregation functions, a powerful and rich family of aggregation functions that have been shown to perform  well for graph representation learning \cite{li2020deepergcn}. The permutation invariant nature of generalized means makes them satisfy Property 1. Further, we show that with appropriate combinations of the intra-edge and inter-edge aggregations, Property 2 is also satisfied. 

Mathematically, generalized means can be expressed as \mbox{$M_p = \left(\frac{1}{n}\sum_{i=1}^{n} x_i^p\right)^\frac{1}{p}$}, where $n$ refers to the number of elements to aggregate, and $p$ denotes its power. The choice of $p$ allows providing different interpretations to the aggregation function. For example, $p=1$ denotes arithmetic mean aggregation, $p=2$ refers to mean squared estimate and a large value of $p$ approximates max pooling from the group. Similarly, $M_p$ can be used for approximating geometric means with $p \rightarrow 0$. We use generalized means for intra-edge as well as inter-edge aggregation. The two functions $\mathcal{F}_1(\cdot)$ and $\mathcal{F}_2(\cdot)$ as stated in Section \ref{sec_msgpass}, for aggregation at node $v_i$ are then defined as

\begin{align}
    & \mathcal{F}_1^{(\mathbf{e})}(\mathbf{s}_1) = \left(  \frac{1}{|\mathcal{N}(v_i)|} \sum_{v_j \in \mathcal{N}(v_i,\mathbf{e})}  w_j \mathbf{x}_{j}^p\right)^{\frac{1}{p}},  \label{eq_agg_mp1}
    \\
    & \text{where}, w_j = \frac{1}{|\mathcal{N}(v_i,\mathbf{e})|} * \left(\sum_{m=1}^{|E(v_i)|}{\frac{1}{|\mathcal{N}(v_i,\mathbf{e}_m)|}}\right)^{-1}, \nonumber \\
    & \mathcal{F}_2(\mathbf{s}_2) = \left(  \frac{1}{|E(v_i)|} \sum_{\mathbf{e}\in E(v_i)}  {\mathbf{s}_2}^p\right)^{\frac{1}{p}}.
    \label{eq_agg_mp2}
\end{align}

For Property 2 to hold in in Eq. \ref{eq_agg_mp1} and Eq. \ref {eq_agg_mp2} above, power term $p$ needs to be the same for $\mathcal{F}_1$ and $\mathcal{F}_2$. Also, the scaling term $w_j$ needs to be added to balance the bias in the weighting introduced in intra-edge aggregation due to varying cardinality across the hyperedges. The related mathematical proof is presented in the Appendix.
 


\subsection{Learning the Importance of Nodes}
Commonly, the contributions from the neighboring nodes within a hyperedge are weighted equally through the choice of simple aggregation functions such as \texttt{mean} and \texttt{max}, among others. Thus, the importance of any node is defined by only its functional characteristics - the values that are contained in its feature vector. However, hypergraphs also possess a complex structure, and this component is almost unused in the learning process. 

We hypothesize that analyzing the structural information can reveal the importance of each node and enhance the message passing process. The structural information of a node $v$ in a hypergraph can be primarily defined by two terms: its global neighbourhood set $\mathcal{N}(v)$ and its hyperedge set $E(v)$. To quantify this information for a node in any graph, \cite{freeman1978centrality} introduced a degree centrality measure which is simply the total number of edges incident on that node. However, compared to graphs where $|\mathcal{N}(v)| = |E(v)|$, these two terms can be significantly different for a hypergraph depending on its structure. Defining empirical degree centrality functions to measure the importance of any node in hypergraphs was shown to be effective, however, such functions are sensitive to the choice of dataset and task \cite{kapoor2013weighted}. 

We introduce a learnable function to quantify the importance of a node which is expressed in terms of $\mathcal{N}$ and $E$ as $\mathcal{C}(\mathcal{N}, E)$. For hypergraphs, node importance function $\mathcal{C}(\mathcal{N}, E)$  is a complex non-linear mapping. Thus, rather than empirically choosing a single  function, we learn the importance value for each node using a small fully-connected neural network comprising two hidden layers. 
For any node $v$, the output of $\mathcal{C}(\cdot, \cdot)$, denoted by $C$, can be interpreted as an attention weight that defines the importance of that node in the message passing process (see Fig. \ref{schematic}). 
Let $C_j$ denote the attention weight for node $v_j \in \mathcal{N}(v_i, \textit{e})$. We conjecture that using $C$ as a weighting term for each node during message passing can improve the learned node embeddings in a hypergraph irrespective of dataset or task.

\subsection{Inductive Learning on Hypergraphs}
Inductive learning of nodes is a challenging problem for hypergraphs as it requires the model to generalize on previously unseen nodes with a diverse set of features and associated sub-hypergraphs. 
Most existing approaches are inherently transductive as they make predictions on nodes in a single, fixed hypergraph. These approaches directly optimize the node representations using matrix-factorization-based objectives, and thus do not generalize to unseen data. 
HyperMSG tackles this challenge through the use of a neural message passing framework as described in section 4.1 that learns to generate embeddings by sampling and aggregating features from a node’s local neighborhood. Our approach uses a neural network comprising $L$ layers, and feature-aggregation is performed at each of these layers, as well as across the hyperedges. Algorithm \ref{algo1} describes the forward propagation mechanism, which implements the aggregation function introduced above. At each iteration, nodes first aggregate information from their neighbors within a specific hyperedge. This is repeated over all the hyperedges across all the $L$ layers. The trainable weight matrices $\mathbf{W}^l$ with $l \in L$ are used to aggregate information across  the feature dimension and propagate it through the hypergraph. The representation on any unseen node can then be obtained by an aggregation process similar to that on seen nodes. 

\begin{algorithm}[]
\SetAlgoLined
\SetKwInOut{Input}{Input}
\SetKwInOut{Output}{Output}
\Input{$\mathcal{H}=(\mathcal{V}, \mathcal{E}, \mathbf{X})$; depth $L$; weight matrices $\mathbf{W}^l$ for $l = 1 \hdots L$; non-linearity $\mathbf{\sigma}$; intra-edge aggregation function $\mathcal{F}_1(\cdot)$; inter-edge aggregation function $\mathcal{F}_2(\cdot)$; node importance function $\mathcal{C}(\cdot, \cdot)$}
\Output{Node embeddings $\mathbf{z}_i | \enskip v_i \in \mathcal{V}$}
$\mathbf{h}_i^0 \leftarrow \mathbf{x}_i \in \mathbf{X} \enskip | \enskip v_i \in \mathcal{V}$

\For {$l = 1 \hdots L$}
{
    $\mathbf{h}_i^l \leftarrow \mathbf{h}_i^{l-1} \enskip ; \enskip \mathbf{s}_2 = \{\emptyset\} $
    
\For {$\mathbf{e} \in E(v_i)$}{ 
    $\mathbf{s}_1 \leftarrow \{\mathcal{C}(\mathcal{N}(v_j), E(v_j)) \odot \mathbf{x}_{j,l-1} \enskip | \enskip v_j \in \mathbf{e}\}$ 
    
    $\mathbf{s}_2 \leftarrow \mathcal{F}_1(\mathbf{s}_1)$
            
     }
     $\mathbf{h}_i^l \leftarrow \mathbf{h}_i^l + \mathcal{F}_2(\mathbf{s}_2)$
     
    $\mathbf{h}_i^l \leftarrow \sigma(\mathbf{W}^l(\mathbf{h}_i^l/||\mathbf{h}_i^l||_2))$
 }
 $\mathbf{z}_i \leftarrow \mathbf{h}_i^L \enskip | \enskip v_i \in \mathcal{V} $
  \caption{HyperMSG Inductive Message Passing}
  \label{algo1}
\end{algorithm}


\textbf{Probabilistic Sampling-based Aggregation. }The modular framework of HyperMSG provides flexibility in adapting the message aggregation module to fit a desired computational memory. This is achieved through aggregating information from only a condensed neighborhood set (Definition 4) $\mathcal{N}{(v_i,\mathbf{e};\alpha)}$ instead of the full neighborhood $\mathcal{N}{(v_i,\mathbf{e})}$.  We propose to apply sub-sampling only on the nodes from the training set, and use information from the full neighborhood for the test set. The advantages of this are twofold. First, a reduced number of samples per aggregation at training time reduces the memory capacity requirement. Second, similar to dropout \cite{srivastava2014dropout}, it serves to add regularization to the optimization process. Using the full neighborhood on test data avoids randomness in the test predictions, and generates consistent output.

For the sampling process, we choose to perform probabilistic selection of the nodes for aggregation based on their importance in the hypergraph using the learned attention weights as described in section 4.2. The probability $P_j$ that the node $v_j$ gets selected in the sampled subset at any iteration is then given by 
\begin{equation}
    P_j = \frac{C_j}{\sum_{j=0}^{|\mathcal{N}(v_i, \mathbf{e})|}C_j}
\end{equation} implying that the neighbor node judged as more important for $v_i$ should be sampled more often.

\textbf{Time complexity analysis.} The time complexity for training HyperMSG is $\mathcal{O}(T|\mathcal{E}|(1+h(d+c)))$ where, $T$ is the total number of training iterations, $d$ denotes the dimensionality of the input feature vector, $h$ is the number of hidden layers, and  $c$ is the number of classes. Thus, the time complexity of HyperMSG is at par with HGNN \cite{feng2019hypergraph} and HyperGCN \cite{yadati2019hypergcn}, with an added advantage of probabilistic sampling by using $\alpha$ that can reduce the memory constraints while also acting as a regularizer. Additional details are provided in the supplementary material.

\section{Experiments}
\label{sec:experiments}
We perform a variety of experiments to evaluate HyperMSG and compare its performance with other hypergraph based learning methods. Firstly, the performance of HyperMSG is evaluated on the task of semi-supervised node classification in hypergraphs through several experiments on representative benchmark datasets. This experiment is performed to test the node-level representative learning capability of HyperMSG and analyze all its possible variants. The results are compared with the state-of-the-art hypergraph representation learning methods. Secondly, we study the stability of HyperMSG by proposing the task of hypergraph classification on an extremely noisy brain neuroimaging dataset. Finally, to show the efficiency of HyperMSG in performing multimodal learning, we compare its performance with recent hypergraph learning methods on a social multimedia dataset. We further show the advantage of using our inductive framework to perform node classification of previously unseen nodes in the multimedia dataset. 


\subsection{Semi-supervised Node Classification in Academic Network}\label{exp1}

\textbf{Experimental setup. }We use the standard co-citation and co-authorship network datasets: CiteSeer, PubMed, Cora \cite{sen2008collective}, DBLP \cite{nr} and arXiv \cite{clement2019use} for this experiment where the task is to predict the topic to which a document belongs (multi-class classification). The input feature vector $\mathbf{x}_{i}$ corresponds to a bag of words, where ${x}_{i,j}$ $\in$ $\mathbf{x}_{i}$ is the normalized frequency of occurrence of the $j^{th}$ word. 
Further, for all experiments, we use a two-layered neural network. All models are implemented in Pytorch and trained using  Adam optimizer \cite{kingma2014adam}. 
Additional implementation details are provided in the Appendix. 

\begin{figure}%
\begin{tikzpicture}
    \centering
    \node[inner sep=0pt] (russell) at (0,0)
    {\includegraphics[scale=0.26]{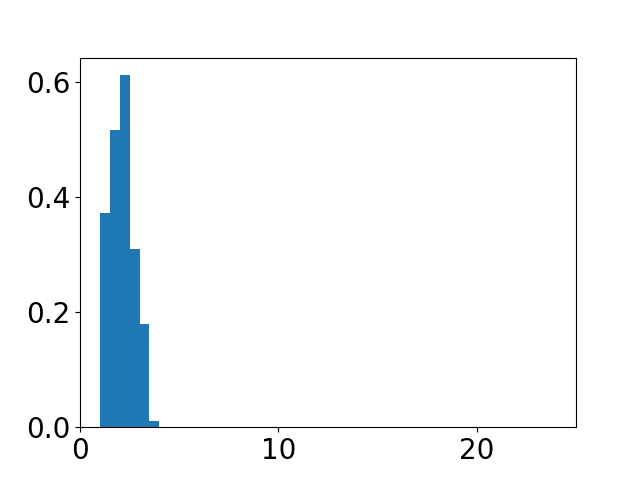}\hspace{-1em}\includegraphics[scale=0.26]{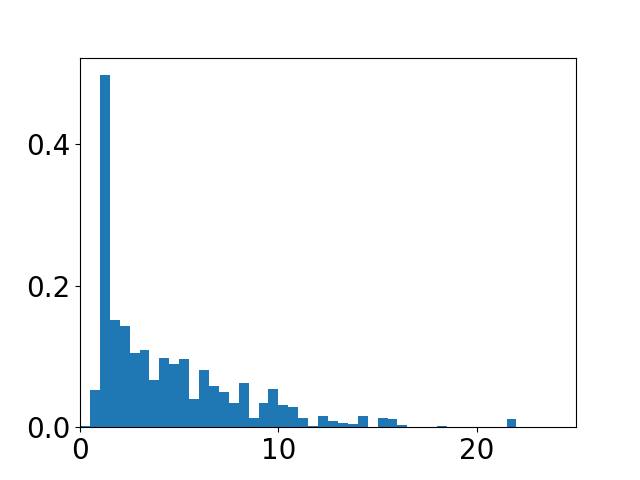}};
    \node[inner sep=0pt, rotate=90] (russell) at (-4.1,0){{\scriptsize Normalized frequency}};
    \node[inner sep=0pt] (russell) at (-1.9,-1.7){{\scriptsize $|\mathcal{N}(v)|/|E(v)|$}};
    \node[inner sep=0pt] (russell) at (1.9,-1.7){{\scriptsize $|\mathcal{N}(v)|/|E(v)|$}};
        \node[inner sep=0pt] (russell) at (1.95,1.4){{\scriptsize CORA co-authorship}};
    \node[inner sep=0pt] (russell) at (-1.9,1.4){{\scriptsize CORA co-citation}};


\end{tikzpicture}
     \caption{Distribution of $|\mathcal{N}(v)|/|E(v)|$ values across CORA co-citation and CORA-co-authorship datasets.}
    \label{cora_analysis}%
\end{figure}
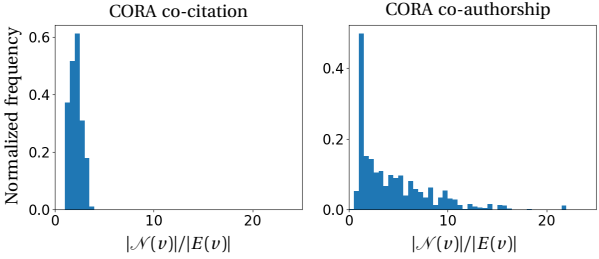

\textbf{Significance of learning the importance of nodes. } For a better understanding on the significance of learning a node importance function for hypergraphs, we show example distributions of  $|\mathcal{N}(v)|/|E(v)|$ for CORA co-citation and co-authorship datasets (see Fig. \ref{cora_analysis}). For the CORA co-citation dataset, $|\mathcal{N}(v)|/|E(v)|$ is mostly close to 1. For this data, we found that the performance gain with message passing using HyperMSG over the graph-based model is relatively small. On the contrary, for CORA co-authorship, the distribution of $|\mathcal{N}(v)|/|E(v)|$ is right-skewed, and has values significantly higher than 1 as well. For this data, HyperMSG outperforms graph-based models for node classification by more than 5\%. This experiment illustrates that even with the same functional characteristics (features) of nodes, the structural information encoded within a hypergraph plays a key role in learning better node embeddings. Additional analysis is presented in the Appendix.

\textbf{Effect of generalized mean aggregations. }We study here the effect of different choices of $p$ in the proposed two-level aggregation function on the performance of HyperMSG. 
\mbox{Fig. \ref{fig_p_val_results}} shows the accuracy scores obtained for 4 different choices of $p$  and $\alpha$ on DBLP and Pubmed datasets. Across all values of $p$, we observe that $p=1$ works best for both datasets. For other choices of $p$, the performance of the model is reduced. For $\alpha=2$, performance of the model seems to be independent of the choice of $p$ for both the datasets. A possible explanation could be that the number of neighbors is very small, and change in $p$ does not affect the propagation of information significantly. 

\pgfplotsset{every axis/.append style={
                    label style={font=\Large},
                    tick label style={font=\Large}  
                    }}

\begin{figure} 
\begin{center}
	\begin{subfigure}{0.47\linewidth}
    \begin{tikzpicture}[scale = 0.45]
        \begin{axis}[%
           title = {{\Large {DBLP}}},
        		thick,
            legend pos = south east,
            xlabel = {Neighborhood Samples ($\alpha$)},
            ylabel = {Accuracy ($\%$)},
            ymin=55, ymax=88,
            log ticks with fixed point,
            xtick={2,4,8,16},
            xmode = log
            ]
            \addplot[mark=*,solid, red, ultra thick] table[x=X,y expr= \thisrow{Y_ours}] {data/data1.dat};
            \addplot[mark=triangle*,solid, blue, ultra thick] table[x=X,y expr = \thisrow{Y_bar}] {data/data1.dat};
            \addplot[mark=square*,solid, brown, ultra thick] table[x=X,y expr = \thisrow{Y_morph}] {data/data1.dat};
            \addplot[mark=diamond*, solid, violet,  ultra thick] table[x=X_lzr,y expr = \thisrow{Y_lzr}] {data/data1.dat};
       \addlegendentry{$p$ = 0.001}
       \addlegendentry{$p$ = 1}
       \addlegendentry{$p$ = 2}
       \addlegendentry{$p$ = 3}
        \end{axis}
    \end{tikzpicture}
   	\label{mto7b}
    \end{subfigure}
	\begin{subfigure}{0.47\linewidth}
    \begin{tikzpicture}[scale = 0.45]
        \begin{axis}[%
           title = {{\Large {Pubmed}}},
        		thick,
            legend pos = south east,
            xlabel = {Neighborhood Samples ($\alpha$)},
            ymin=55, ymax=88,
            log ticks with fixed point,
            xtick={2,4,8,16},
            xmode = log
            ]

            \addplot[mark=*,solid, red, ultra thick] table[x=X,y expr= \thisrow{Y_ours}] {data/data2.dat};
            \addplot[mark=triangle*,solid, blue, ultra thick] table[x=X,y expr = \thisrow{Y_bar}] {data/data2.dat};
            \addplot[mark=square*,solid, brown, ultra thick] table[x=X,y expr = \thisrow{Y_morph}] {data/data2.dat};
            \addplot[mark=diamond*, solid, violet,  ultra thick] table[x=X_lzr,y expr = \thisrow{Y_lzr}] {data/data2.dat};
       \addlegendentry{$p$ = 0.001}
       \addlegendentry{$p$ = 1}
       \addlegendentry{$p$ = 2}
       \addlegendentry{$p$ = 3}
        \end{axis}
    \end{tikzpicture}
   	\label{mto7b}
    \end{subfigure}
\end{center}
\vspace{-1em}
\caption{Performance of HyperMSG for different choices of generalized means ($p$) and neighborhood samples ($\alpha$).}
\label{fig_p_val_results}	
\end{figure}
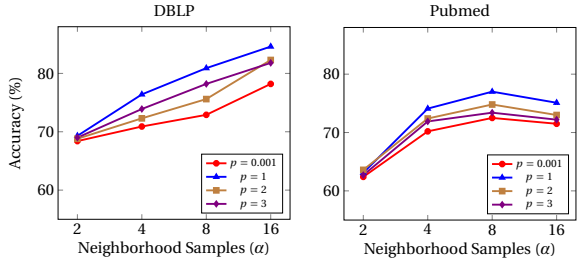

\textbf{Effect of probabilistic sampling. }We study here the effect of number of samples per aggregation on the performance of the model (Fig. \ref{fig_p_val_results}). For DBLP, model performance increases with increasing value of $\alpha$. However, for Pubmed, we observe that performance improves up to $\alpha=8$, but then a slight drop is observed for larger sets of neighbors. Note that for Pubmed, the majority of the hyperedges have cardinality less than or equal to 10. This means that for $\alpha=16$, information will be aggregated from almost all the neighbors, thereby involving almost no random sampling. Stochastic sampling of nodes can serve as a regularization mechanism and reduce the impact of noisy hyperedges. This is possibly the reason that performance for $\alpha=8$ is higher than for 16.

\begin{table*}
\centering\fontsize{9}{11}\selectfont
\caption{Performance scores (in terms of accuracy \%) for various hypergraph learning methods on co-authorship or co-citation datasets. The term `Hom.' denotes homogeneous networks that use either of co-citation and co-authorship datasets, and `Het.' refers to those using the combination of both datasets. }
\vspace{0.1 in}
\begin{tabular}{llcccccc}
\toprule
 
 \textbf{Data} &  \textbf{Method} &Cora &DBLP &arXiv&Pubmed&Citeseer \\
\cmidrule(l){1-1}\cmidrule(l){2-2} \cmidrule(l){3-3} \cmidrule(l){4-4} \cmidrule(l){5-5} \cmidrule(l){6-6} \cmidrule(l){7-7} 

& MLP + HLR \cite{yadati2019hypergcn}   &   63.1 $\pm$ 1.8             &    61.6 $\pm$ 2.1                          &     61.7 $\pm$ 2.3 & 69.1 $\pm$ 1.5 & 62.3  $\pm$ 1.6  \\ [3pt]
  & HGNN \cite{feng2019hypergraph}  &   66.3 $\pm$ 2.8             &    73.8 $\pm$ 2.1                          &     68.1 $\pm$ 2.7 & 68.1 $\pm$ 3.5 & 62.6 $\pm$ 1.6 \\ [3pt]
Hom. &  HyperGCN \cite{yadati2019hypergcn}  &   69.7 $\pm$ 3.7             &    74.2 $\pm$ 5.2                          &     68.2 $\pm$ 3.6 & 73.4 $\pm$ 3.8 & 62.7 $\pm$ 4.6 \\ [3pt]
&  UniGAT \cite{huang2021unignn}  &   75.0 $\pm$ 1.1            &    87.8 $\pm$ 1.1                          &     77.2 $\pm$ 1.3 & 74.6 $\pm$ 1.2 & 63.8 $\pm$ 1.5 \\ [3pt]
&  UniGCN \cite{huang2021unignn}  &   75.3 $\pm$ 1.3             &    \textbf{88.0 $\pm$ 1.1}                          &     77.3 $\pm$ 1.8 & 74.1 $\pm$ 1.0 & 63.7 $\pm$ 1.5 \\ [3pt]
\toprule
\multirow{ 5}{*}{Het.}& HetGNN \cite{zhang2019heterogeneous}  &   72.6 $\pm$ 1.3             &    77.9 $\pm$ 2.0                         &     74.5 $\pm$ 2.0 & - & - \\ [3pt]
& HAN \cite{wang2019heterogeneous}  &   72.8 $\pm$ 1.9             &    77.9 $\pm$ 1.4                          &     75.0 $\pm$ 2.2 & - & - \\ [3pt]
& MAGNN \cite{fu2020magnn}  &   73.3 $\pm$ 1.5             &    78.3 $\pm$ 1.8                        &     75.8 $\pm$ 1.8 & - & - \\ [3pt]
& MPNN-R \cite{yadati2020neural}  &   74.7 $\pm$ 1.5             &    78.6 $\pm$ 1.7                        &     77.7 $\pm$ 1.7 & - & - \\ [3pt]
\toprule
\multirow{ 2}{*}{Hom.}& HyperMSG (non-adaptive, \textbf{ours})   &   74.9 $\pm$ 1.0           &    80.4 $\pm$ 1.1                     &     $78.6 \pm 1.4$ & 76.2 $\pm$ 1.6 & 66.4 $\pm$ 1.8 \\ [3pt]
& HyperMSG (\textbf{ours})   &   \textbf{77.7 $\pm$ 1.2}             &   85.7 $\pm$ 1.1                        &     \textbf{79.1 $\pm$ 1.6} & \textbf{77.1 $\pm$ 1.2} & \textbf{66.8 $\pm$ 1.6} \\ [3pt]
 \bottomrule
\end{tabular}
\label{table_exp1}
\end{table*}

\textbf{Performance comparison with existing methods. }In \mbox{Table \ref{table_exp1}}, we compare the results of HyperMSG with state-of-the-art hypergraph learning methods. Among these, the homogeneous networks use either the co-citation or co-authorship datasets. The heterogeneous networks combine information from both co-citation and co-authorship datasets. For HyperMSG, we use homogeneous network for a fair comparison. We report the results with arithmetic mean $p = 1$ using the complete neighborhood i.e., $\alpha = |\mathbf{e}|$. 
To show the significance of learning importance of nodes, we report the results for HyperMSG as well as its non-adaptive variant in Table \ref{table_exp1}. For all models, 10 data splits over 8 random weight initializations are used, totalling 80 experiments per method for every dataset. The data splits are the same as in HyperGCN and are  described in the Appendix. Note that for Pubmed and Citeseer dataset, the co-authorship information does not exist and hence, the heterogeneous models are not applicable.

From Table \ref{table_exp1}, we observe that both variants of HyperMSG outperform the homogeneous networks by remarkable margins. Interestingly, our implementations of HyperMSG with only homogeneous information are able to outperform the heterogenous networks that use information from two different sources. This clearly demonstrates that HyperMSG exhibits strong representative power and is able to extract information from the hypergraphs at levels beyond the existing baselines. 

\pgfplotsset{every axis/.append style={
                    label style={font=\Large},
                    tick label style={font=\Large}  
                    }}
\begin{figure} 
\centering
	\begin{subfigure}{0.47\linewidth}
\begin{tikzpicture}[scale = 0.45]
  \begin{axis}
  [%
           title = {{\Large {DBLP}}},
        		thick,
            legend pos = south east,
            xlabel = {Train:Test Ratio},
            ylabel = {Accuracy ($\%$)},
            ymin=60, ymax=95,
            xtick={0,1,2,3,4},
    xticklabels={$1/50$,$1/30$,$1/15$,$1/10$,$1/3$}
            ]
  
    \addplot [solid, color = blue, mark=asterisk, error bars/.cd, y dir=both, y explicit,
      error bar style={line width=2pt,solid},
      error mark options={line width=1pt,mark size=4pt,rotate=90}]
    table [x=x, y=y, y error=y-err]{%
      x y y-err
      0 84.7 2.3
      1 85.9 2.1
      2 86.3 1.8
      3 86.9 1.4
      4 87.2 1.1
    };
    \addplot [solid,color = red, mark=asterisk, error bars/.cd, y dir=both, y explicit,
      error bar style={line width=2pt,solid, color = red},
      error mark options={line width=1pt,mark size=4pt,rotate=90}]
    table [x=x, y=y, y error=y-err]{%
      x y y-err
      0 73.1 6.4
      1 75.3 5.2
      2 77.3 4.3
      3 79.9 3.3
      4 80.9 3.0
    };
    \addlegendentry{HyperMSG}
  \addlegendentry{HyperGCN}

  \end{axis}
 \end{tikzpicture}
 \end{subfigure}
 \hspace{-0.2cm}
\begin{subfigure}{0.47\linewidth}
\begin{tikzpicture}[scale = 0.45]
  \begin{axis}
  [%
           title = {{\Large {Pubmed}}},
        		thick,
            legend pos = south east,
            xlabel = {Train:Test Ratio},
            ymin=60, ymax=95,
    xtick={0,1,2,3,4},
    xticklabels={$1/150$,$1/100$,$1/50$,$1/3$, $1/3$},
            ]
    \addplot [solid, color = blue,mark=asterisk, error bars/.cd, y dir=both, y explicit,
      error bar style={line width=2pt,solid},
      error mark options={line width=1pt,mark size=4pt,rotate=90}]
    table [x=x, y=y, y error=y-err]{%
      x y y-err
      0 74.3 2.5
      1 76.9 2.2
      2 79.2 1.6
      3 80.1 1.3
      4 80.5 1.0
    };
    \addplot [solid,color = red, mark=asterisk, error bars/.cd, y dir=both, y explicit,
      error bar style={line width=2pt,solid, color = red},
      error mark options={line width=1pt,mark size=4pt,rotate=90}]
    table [x=x, y=y, y error=y-err]{%
      x y y-err
      0 70.8 6.1
      1 73.8 5.4
      2 75.0 5.0
      3 75.5 4.5
      4 76.9 4.0
    };
            \addlegendentry{HyperMSG}
       \addlegendentry{HyperGCN}
  \end{axis}
 \end{tikzpicture}
 \end{subfigure}
     \caption{Accuracy scores for HyperMSG and HyperGCN obtained for different train-test ratios. }
    \label{fig_stable}%
 \end{figure}
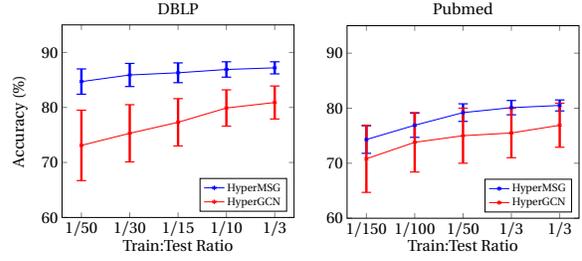

\textbf{Stability analysis. }We further study the stability of our method in terms of the variance observed in performance for different train-test split ratios. We compare the results with HyperGCN under similar settings, as it is the state-of-the-art method of a broad set of hypergraph learning methods which are based on some sort of hypergraph to graph conversion. Fig. \ref{fig_stable} shows results for the two learning methods on 5 different train-test ratios. We see that the performance of both models improves when a higher fraction of data is used for training, and the difference in their performances decreases at the train-test ratio of 1/3.
However, for smaller ratios, we see that HyperMSG outperforms HyperGCN across all datasets. Further, the standard deviation for the predictions of HyperMSG is  lower than that of HyperGCN. Clearly, this implies that HyperMSG is able to better exploit the information contained in the hypergraph compared to HyperGCN, and can thus produce more accurate and stable predictions. 
Results of the experiment on Cora and Citeseer, which exhibit a similar trend, can be found in the Appendix.


\begin{table*}
\centering\fontsize{9}{11}\selectfont
\caption{Performance of HyperMSG and its variants on nodes which were part of the training hypergraph (seen) and nodes which were not part of the training hypergraph (unseen).}
\vspace{0.1 in}
\begin{tabular}{lcccccccc}
\toprule
  &\multicolumn{2}{c}{DBLP} & \multicolumn{2}{c}{Pubmed} &\multicolumn{2}{c}{Citeseer} & \multicolumn{2}{c}{Cora (citation)}\\ 
 \cmidrule(lr){2-3} \cmidrule(lr){4-5} \cmidrule(lr){6-7} \cmidrule(lr){8-9}
  \textbf{Method} &Seen &Unseen &Seen &Unseen &Seen &Unseen &Seen &Unseen\\
 \cmidrule(l){2-2} \cmidrule(l){3-3} \cmidrule(l){4-4} \cmidrule(l){5-5}  \cmidrule(l){6-6} \cmidrule(l){7-7} \cmidrule(l){8-8} \cmidrule(l){9-9}
 
MLP + HLR  & 64.5$\pm$2.5 &58.7$\pm$3.1 &66.8$\pm$2.4 &62.4$\pm$3.5 &60.1$\pm$1.2 &58.2$\pm$1.9 &65.7$\pm$2.3 &64.2$\pm$2.5  \\
UniGCN & \textbf{88.5$\pm$1.}2 & \textbf{82.6$\pm$2.2} &83.7$\pm$1.1 &83.3$\pm$1.3 &71.2$\pm$1.2 &\textbf{70.6$\pm$1.9} &74.3$\pm$2.3 &71.5$\pm$2.5  \\
\toprule
HyperMSG ($\alpha$ = 4) & 84.7$\pm$2.8 & 72.2$\pm$2.9 &79.2 $\pm$ 3.1 &70.4 $\pm$ 2.6  & 69.3$\pm$1.9&68.8$\pm$2.9 & 74.8 $\pm$ 1.5 & 73.2 $\pm$ 2.0 \\
HyperMSG ($\alpha$ = 8) & 87.5$\pm$1.9 & 77.7$\pm$1.1 &\textbf{84.4$\pm$1.4} & \textbf{83.5$\pm$1.2}  & 70.8$\pm$1.8&69.6$\pm$1.6 & \textbf{75.4 $\pm$ 0.8} & 74.6 $\pm$ 1.7 \\
HyperMSG ($\alpha$ = 16) &88.0$\pm$1.3 & 81.5$\pm$ 1.1 &83.6$\pm$0.8 & 81.7$\pm$0.9  & \textbf{71.7$\pm$0.8}& \textbf{70.6$\pm$1.0} & 75.1 $\pm$ 0.5 & \textbf{74.8 $\pm$ 1.3} \\
 
\bottomrule
\end{tabular}

\label{tab_inductive}

\end{table*}

\textbf{Classification of unseen nodes. }To test the performance of HyperMSG on unseen nodes, we create inductive learning datasets for DBLP, Pubmed and Citeseer. To do so, we split each dataset into a ratio of 1:3:1 for the training, validation (seen) and test (unseen) sets, respectively. To obtain the unseen test set, we break the hypergraph into two sub-hypergraphs. Note that this splitting leads to several node connections being disregarded as well as a relatively sparse test hypergraph. This can induce noise in the learning process, which we tackle by employing our probabilistic sampling mechanism in message passing. We show that by capturing all the structural and functional properties of a hypergraph, HyperMSG performs better than the other models.

Table \ref{tab_inductive} presents the performance scores of HyperMSG for different choices of $\alpha$. To the best of our knowledge, no competitive baseline inductive learning method existed before our preliminary work HyperSAGE \cite{arya2020hypersage} which introduced a two-level message aggregation framework. Recently, \cite{huang2021unignn} further extended the two-level message aggregation framework and proposed a set of inductive learning methods. In this experiment, we compare the performance of HyperMSG with the base MLP+HLR approach and UniGCN \cite{huang2021unignn} as a reference.  A general observation is that HyperMSG works well for unseen nodes and significantly outperforms the other models except in the case of DBLP. Further, the difference between performance scores for seen and unseen nodes is stable across all datasets. Based on these observations, it can be concluded that HyperMSG works well in an inductive setting, i.e. on unseen nodes, as well.


\subsection{Hypergraph classification on neuroimaging data}
\begin{figure}
    \centering
    \includegraphics[scale =0.45]{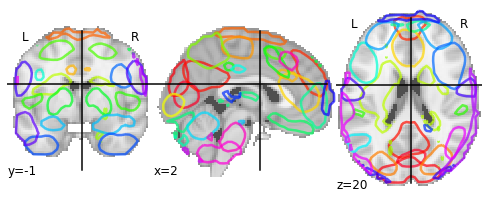}
    \caption{Example slices of a brain sample from the autism neuroimaging data \protect\cite{craddock2013neuro} for the three planes showing the construction of hypergraph.}
    \label{fig_brain}
\end{figure}

\begin{table}
\centering\fontsize{9}{11}\selectfont
\caption{Performance scores for HyperMSG and  baseline hypergraph learning methods for the task of brain classification for autism spectral disorder based on neuroimaging data.}
\vspace{0.1 in}
\begin{tabular}{lc}
\toprule
Method & AUC-ROC\\
 \cmidrule(l){1-1} \cmidrule(l){2-2} 
 
HGNN \cite{feng2019hypergraph} & 62.5 $\pm$ 4.9  \\
HyperGCN \cite{yadati2019hypergcn} & 59.2 $\pm$ 5.8  \\
HyperMSG (non-adaptive, \textbf{ours}) & 64.3 $\pm$ 4.4  \\
HyperMSG (\textbf{ours}) & \textbf{67.2 $\pm$ 5.3}  \\
 
\bottomrule
\end{tabular}
\label{tab_braindata}

\end{table}

 To demonstrate the generality of HyperMSG, we use it on the task of hypergraph classification. For performing hypergraph classification, the model needs to integrate the learned features from each node in a way that the combined features are significantly dissimilar across  hypergraphs. This makes hypergraph classification a challenging task as compared to the previous node classification tasks. Further, to also study the robustness of the proposed model under noisy scenarios, we choose an extremely noisy brain neuroimaging dataset \cite{craddock2013neuro}, where each brain image is a hypergraph. It involves classification of control subjects for autism spectrum disorder (ASD) using 4D resting-state functional magnetic resonance imaging (fMRI) data. 
 Typically, an fMRI sample comprises about 20,000 voxels with 300 time points, making it extremely high-dimensional with a significant level of inherent noise \cite{parisot2017flexible, esteban2017mriqc}. 
 
 Some earlier methods have sought to mitigate this problem through aggregating information along one of the dimensions, thus leading to a 3D volume \cite{el2019simple, thomas2020classifying}. Other approaches involve summarization of the data as cross-correlation matrix between macro regions of the brain \cite{parisot2018disease,arya2020fusing}. Although these solutions simplify data handling, they come at the expense of significant information loss. Hypergraphs can be used to handle this task without any such approximations.
An example representation of a brain hypergraph is shown in Fig. \ref{fig_brain}. More details about the processing steps and data preparation are provided in the Appendix.

 Table \ref{tab_braindata} shows the performance scores for the various hypergraph-based methods. Due to class imbalance in the dataset, we use AUC-ROC to measure the performance. We see that both variants of HyperMSG outperform HGNN, HyperGCN and UniGCN methods, with our adaptive variant showing an improvement of around 5\% over HGNN. The results show the robustness and stability of HyperMSG in utilizing the information within hypergraphs. The large variance observed in performance for all the methods can be attributed to the low signal-to-noise ratio of the dataset.




\subsection{Multimodal Hypergraph Analysis}

\begin{figure}
    \centering
    \includegraphics[scale=0.25]{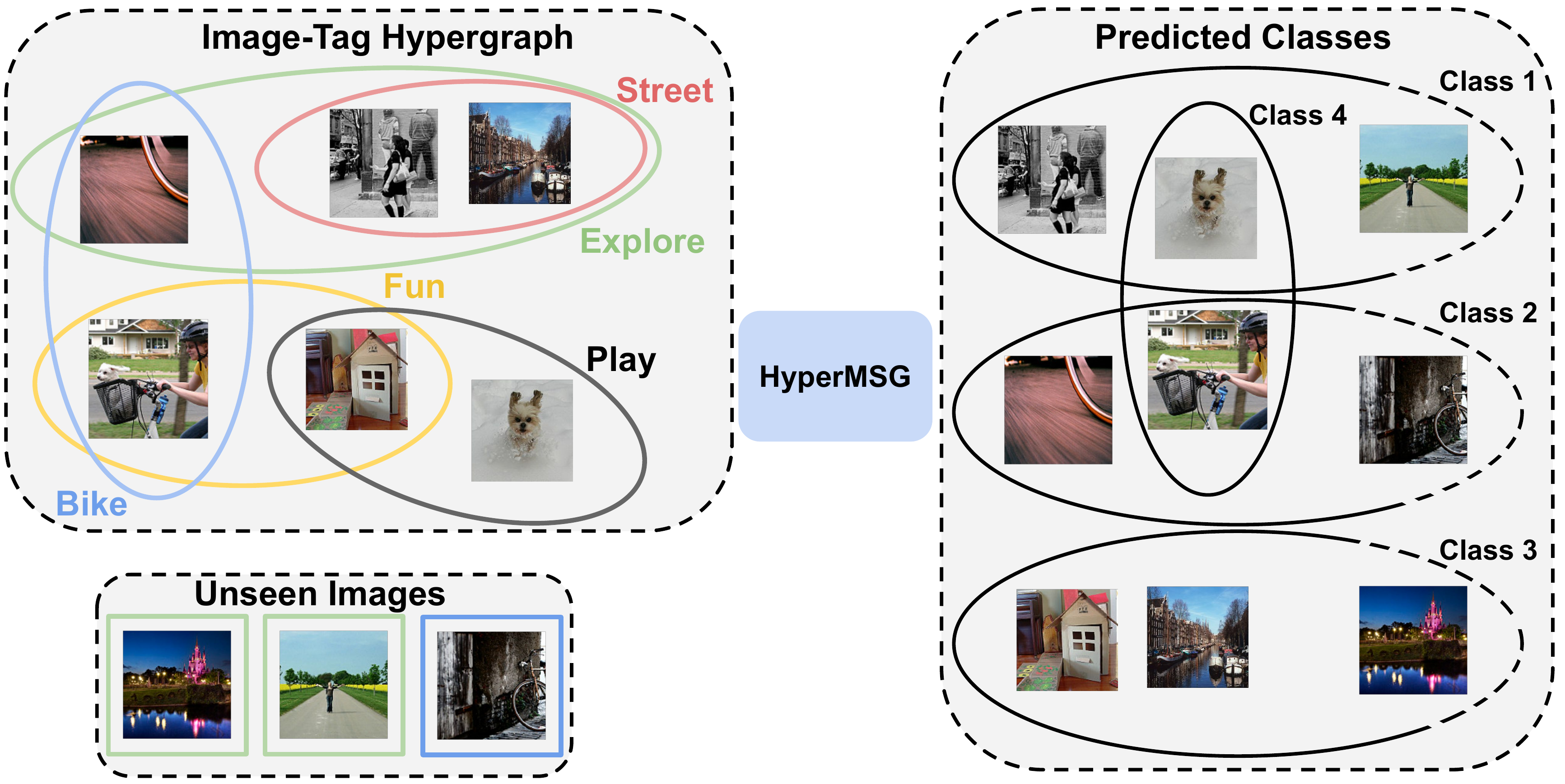}
    \caption{An example of the proposed multimodal hypergraph analysis on the Flickr dataset. The input to the model is an image-tag hypergraph (left) where each image is represented on a node and the hyperedges correspond to the tags associated with the images.
    The goal is to classify an image into its respective class which can be the ground truth \textit{labels} in case of \textit{Task 1}, \textit{users} in case of \textit{Task 2} and \textit{groups} in case of \textit{Task 3}. Unlike previous methods such as \cite{arya2018exploiting}, HyperMSG can perform inferences on unseen images as well.}
    \label{multimodal_schematic}
\end{figure}

Social networks are rich with multimodal information and learning an effective representation for the entities of interest in them, such as users, images and text, has gained a great attention in different applications.
Examples include node classification \cite{bhagat2011node, mcauley2012image}, link prediction \cite{liben2007link, li2013link}, community detection \cite{leskovec2010empirical, arya2019predicting}, and network visualization \cite{tang2016visualizing, fischer2020visual}.
In this section, we evaluate the performance of HyperMSG on MIR Flickr \cite{huiskes2008mir}, a social multimedia dataset commonly used for multimodal learning on hypergraphs \cite{xu2012unified, arya2018exploiting, arya2019hyperlearn}.

\textbf{Experimental setup. } The MIR Flickr dataset contains heterogeneous entities and relationships among them. In particular, the  dataset consists of 25,000 images ($I$) from Flickr posted by 6,386 users ($U$) to 10,575 groups ($G$) and annotated with over 50,000 user-provided tags ($T$). The dataset also provides manually-created ground truth image annotations at the semantic concept level. There are 25 such unique concepts in the dataset. Similar to the related multimedia works \cite{li2016weakly,arya2019hyperlearn}, we remove noisy tags occurring less than 50 times and obtain a vocabulary consisting of 457 tags. Accordingly, 18 concepts are preserved and utilized to validate the performance in our experiments. 
In this work, we follow \cite{arya2018exploiting} as a testbed for demonstrating multimodal learning capabilities of HyperMSG by performing 3 types of tasks:
\textit{Task 1}: Multi-Label Image Classification, \textit{Task 2}: Image-User Link Prediction and \textit{Task 3}: Group Recommendation.


\begin{figure*}
    \centering
    \includegraphics[height=5.3cm, width = \textwidth]{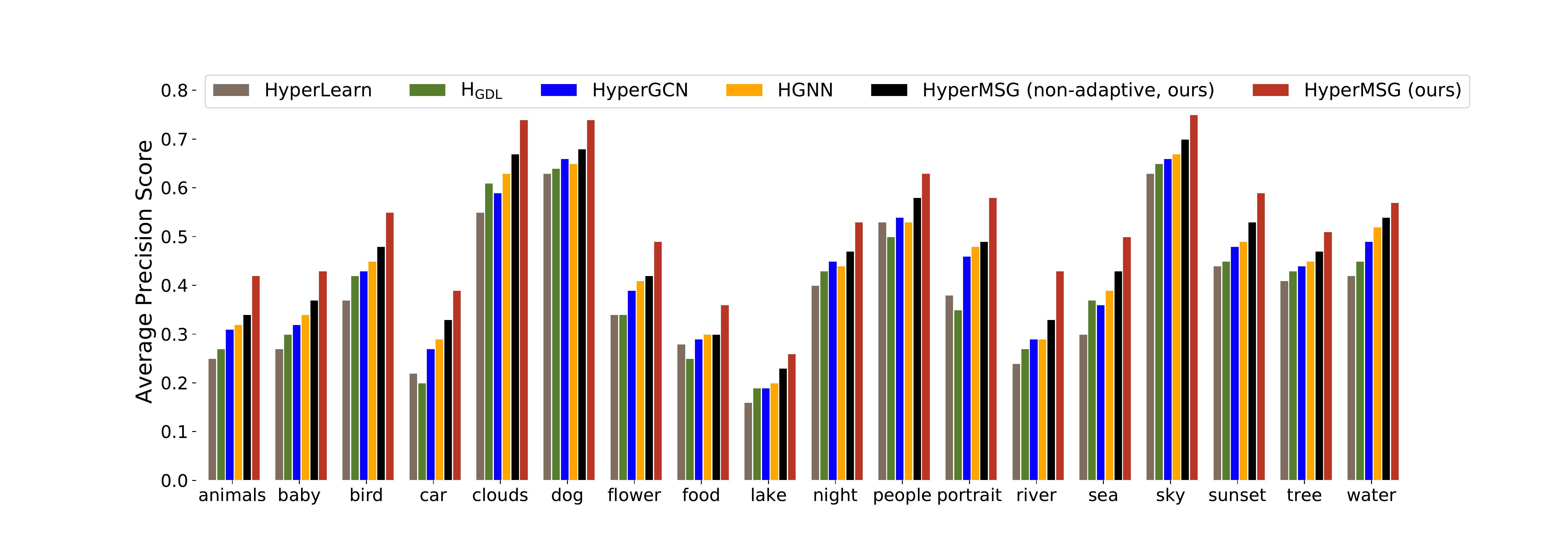}
    \caption{\textit{Task 1} (Multi-Label Image Classification) Detailed performance comparison in terms of Average Precision over 18 concepts on the MIRFlickr dataset}
    \label{concept_result}
\end{figure*}

\begin{figure}
    \centering
    \begin{subfigure}[t]{0.48\linewidth}
        \centering
        \includegraphics[height=1.5in]{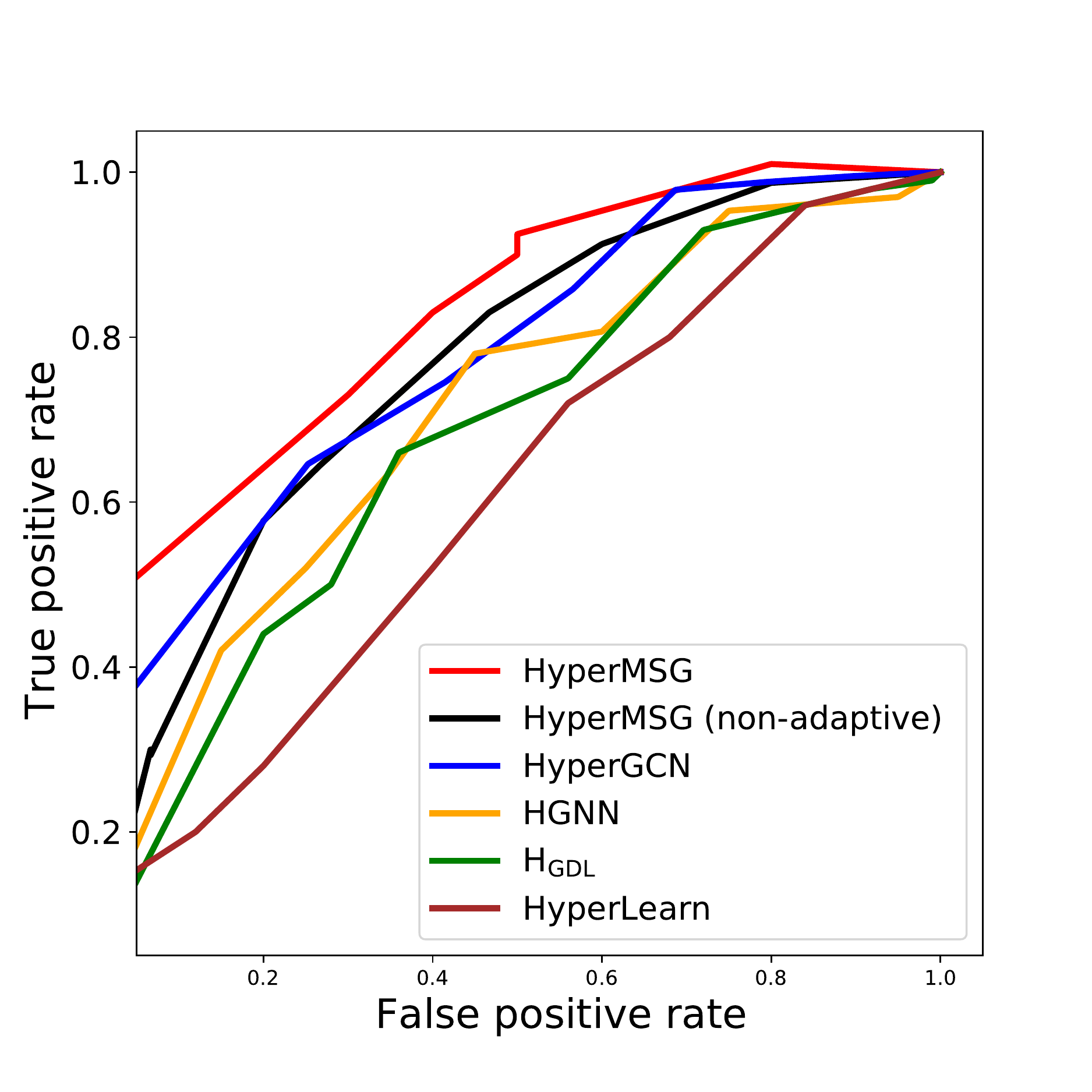}
        \caption{\textit{Task 2}: Image-User Link Prediction}
    \end{subfigure}%
    ~ 
    \begin{subfigure}[t]{0.48\linewidth}
        \centering
        \includegraphics[height=1.5in]{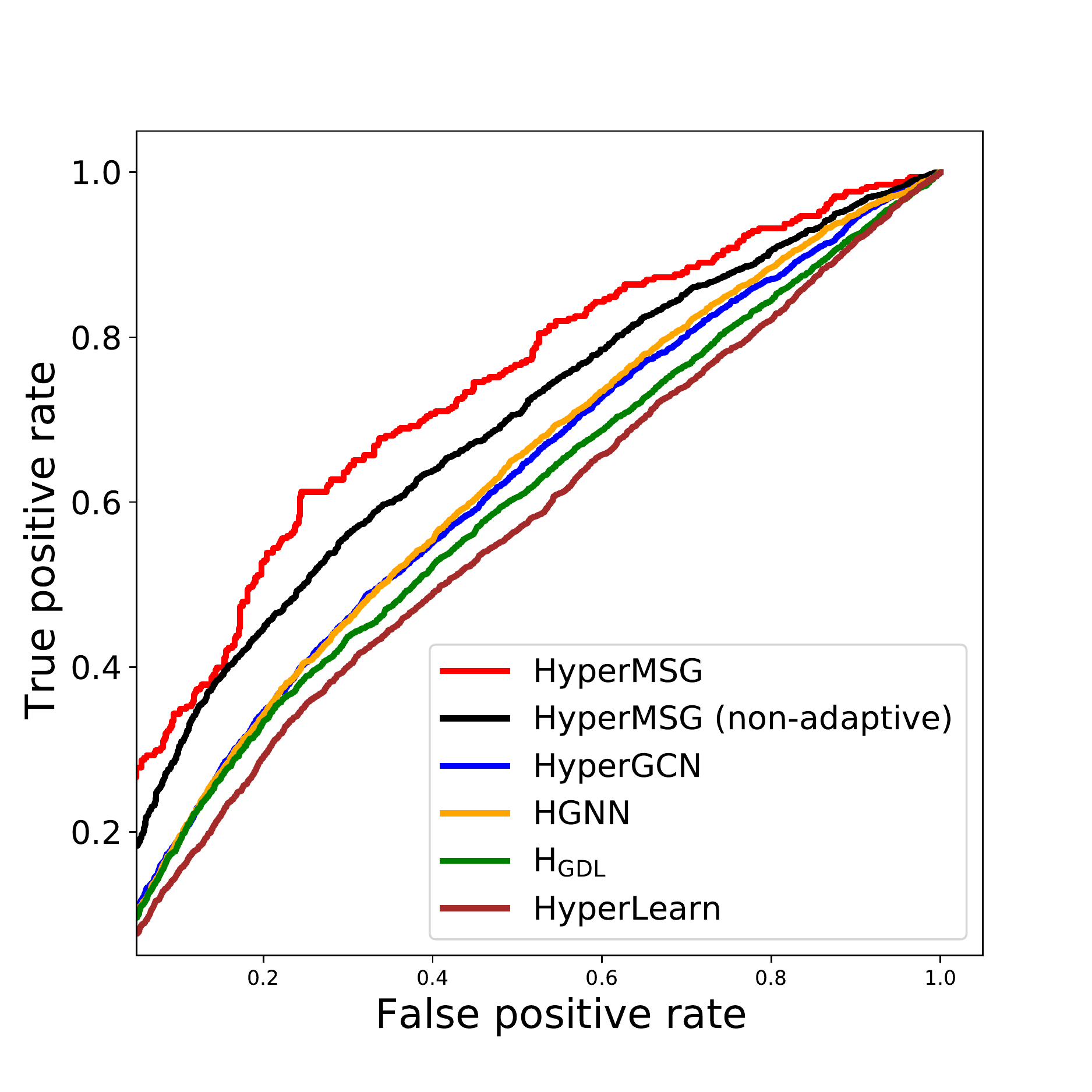}
        \caption{\textit{Task 3}: Group Recommendation}
    \end{subfigure}
    \caption{Receiver Operating Characteristics (ROC) curve showing the performance of the models on (a) Image-User Link Prediction and (b) Group Recommendation. On both tasks HyperMSG consistently outperforms alternative multimodal hypergraph learning methods.}
    \label{roc_plot_multimodal}
\end{figure}

\textbf{Hypergraph Construction} We create a hypergraph where each node represents an image and the hyperedges (relations between images) are formed using the tags, similar to our previous work \cite{arya2018exploiting}. For all the three tasks hyperedges correspond to the image-tag relationships i.e. a hyperedge is constructed among images sharing a common tag. Subsequently, each image is associated with ground truth labels, users and groups respectively for the corresponding three tasks. Figure \ref{multimodal_schematic} shows the image-tag hypergraph (left) and the predicted classes (right) i.e. the ground-truth labels in case of \textit{Task 1}, users in case of \textit{Task 2} and groups in case of \textit{Task 3}.

\textbf{Performance comparison with existing methods. } 
We compare the performance of our proposed HyperMSG model with the following hypergraph learning methods: HyperGCN \cite{yadati2019hypergcn}, HGNN \cite{feng2019hypergraph}, HyperLearn \cite{arya2019hyperlearn} and $H_{GDL}$ \cite{arya2018exploiting}. For a fair and thorough comparison, we conduct the experiments using only structural information and hence we assign identity features to each image (node of the hypergraph). In the case of \textit{Task 1}, we provide a detailed performance analysis on each of the concepts. We report the performance of our proposed HyperMSG approach and the alternatives in terms of Average Precision (AP), a standard evaluation measure used for multi-label image classification benchmarks. For the other two tasks, we combine the classification scores for all the users/groups and plot their respective Receiver Operating Characteristics (ROC) curves. The ROC curve depicts how well a model is able to predict the presence/absence of a class among images with the corresponding metadata. 
Figure~\ref{concept_result} shows that both variants of HyperMSG consistently outperform the other hypergraph learning methods in \textit{Task 1}, i.e. classifying images into the 18 concepts.
The results of this experiment suggest that our proposed approach is more effective in modeling social image-tag relationships than the state-of-the-art alternatives. 
The advantage of HyperMSG in learning multimodal relations as compared to the other hypergraph methods is further confirmed by the ROC curves shown in Figure~\ref{roc_plot_multimodal} for \textit{Task 2} and \textit{Task 3}. In all three experiments we observe that the adaptive version of HyperMSG utilizing attention-based mechanism comes out as the best performer. 
We hypothesize that this can be attributed to the significance of global and local neighborhoods in social multimedia networks where the nodes that have many connections (high-degree nodes) tend to be connected to the other nodes with many connections, while they are surrounded by many small clusters of low-degree nodes \cite{mislove2007measurement}. 

\textbf{Inductive Learning.} To evaluate the performance of HyperMSG on unseen nodes, we perform multi-label image classification on the MIR Flickr dataset. For this experiment, we include the features of the images as well, which we extract from the penultimate layer of a pre-trained VGG-16 network \cite{simonyan2014very}. Similar to Section~\ref {exp1}, we compare our method with inductive learning methods, MLP+HLR and UniGCN and report the average precision (AP) scores averaged over all the classes. We split the dataset into the training, validation (seen) and test (unseen) samples according to a $1:3:1$ ratio. To obtain the unseen test set, we break the hypergraph into two sub-hypergraphs, ensuring that the training set contains at least one image from each class.
Table~\ref{tab_inductive_multimodal} shows that the HyperMSG performs better than the alternatives on both seen and unseen nodes, while also producing a lower standard deviation in the performance level.

\begin{table}
\centering\fontsize{9}{11}\selectfont
\caption{Average precision of  HyperMSG and baseline hypergraph learning methods for the task of multi-label image classification on MIR Flickr dataset on both seen and unseen nodes.}
\vspace{0.1 in}
\begin{tabular}{lcc}
\toprule
Method & Seen & Unseen\\
 \cmidrule(l){1-1} \cmidrule(l){2-2} \cmidrule(l){3-3} 
 
MLP + HLR  & 0.64 $\pm$ 1.3 & 0.61 $\pm$ 1.4\\
UniGCN  & 0.70 $\pm$ 1.1  & 0.67 $\pm$ 1.3 \\
HyperMSG (non-adaptive, \textbf{ours}) & 0.70 $\pm$ 1.1 & 0.68 $\pm$ 1.3 \\
HyperMSG (\textbf{ours}) & \textbf{0.73 $\pm$ 0.9}  & \textbf{0.69 $\pm$ 1.1} \\
\bottomrule
\end{tabular}
\label{tab_inductive_multimodal}

\end{table}

\section{Conclusion}
\label{sec:conclusion}
In this paper, we have presented HyperMSG, a two-level neural message passing framework for inductive learning on hypergraphs. HyperMSG fully utilizes the inherent higher-order relations in a hypergraph structure without reducing it to an intermediate graph representation. It 
adaptively learns the importance of each node during the learning process, thereby improving the message passing process. Through experiments on several representative datasets, we have shown that HyperMSG outperforms the existing methods for hypergraph learning.
We have demonstrated that HyperMSG generates stable predictions for very sparse sampling as well as when the nodes are unseen during the training process.  From the results on the challenging task of brain image classification for autism,  we conclude that HyperMSG yields accurate and robust results in hypergraph classification even on extremely noisy time-series datasets. Finally, HyperMSG is suitable for performing mutimodal representation learning tasks and can outperform all existing hypergraph based learning methods on information rich multimedia data.

\ifCLASSOPTIONcompsoc
  \section*{Acknowledgments}
\else
  \section*{Acknowledgment}
\fi

This research has received funding from the European Union’s
Horizon 2020 research and innovation programme under grant
agreement 700381 (ASGARD project). The authors would like to thank Selene Gallo, Dr. Rajat Thomas and their team at Amsterdam Medical Center for their invaluable support in pre-processing the neuroimaging dataset.

\ifCLASSOPTIONcaptionsoff
  \newpage
\fi

\bibliographystyle{plain}
\bibliography{citations}

\onecolumn

\section*{Appendix}
\label{sec:appendix}
\appendices
\section{Clique Expansion on Fano Planes}
\begin{figure}[!htb]
\centering
    \begin{subfigure}{0.5\textwidth}
    \includegraphics[width=8cm]{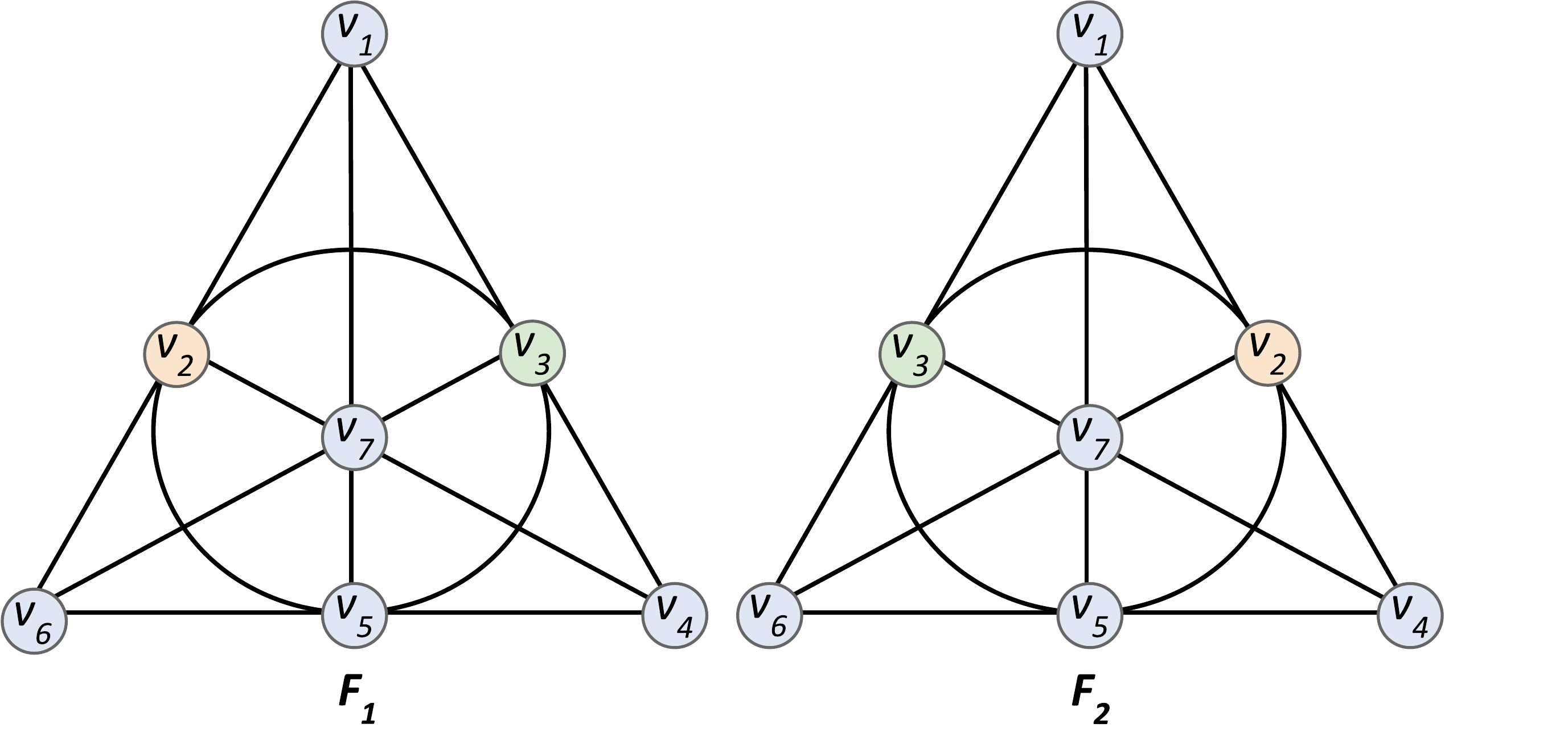}%
    \end{subfigure}
     \caption{The Fano plane ($F_1$), and a copy ($F_2)$ with the nodes $v_2$ and $v_3$ permuted}
    \label{fig_fano_suppl}%
\end{figure}

Converting a hypergraph to graph using methods based on clique or star expansions can lead to partial loss of the inherent structural information. Examples highlighting this for clique, star and functional have been presented in Section 1 of the main paper. We present here another example of two distinct Fano planes $F_1$ and $F_2$, see \mbox{Fig. \ref{fig_fano_suppl}}. Both comprise 7 nodes $\{{v_1, v_2, v_3, v_4, v_5, v_6, v_7}\}$. Fano plane $F_1$ has 7 hyperedges $\{\{v_1, v_2, v_6\}, \{v_1, v_3, v_4\}, \{v_1, v_5, v_7\},\{v_2, v_3, v_5\}, \{v_2, v_4, v_7\}, \{v_3, v_6, v_7\}$, $\{v_4, v_5, v_6\}\}$ (6 straight lines and 1 circle) and $F_2$ is a copy of the first with nodes $v_2$ and $v_3$ permuted. Since each pair of distinct nodes lies in exactly one hyperedge, the clique expansion for $F_1$ and $F_2$ will be will be a complete graph of 7 nodes (K7), thus identical. In fact, there can be 30 such Fano planes obtained by permuting nodes, whose clique expansion will be identical \cite{dong2020hnhn}. 
Hence, converting hypergraph to graph and applying graph-based message passing will learn sub-optimal representations and lose structural information which is important to distinguish between the two hypergraphs. 

\section{Global Neighborhood Invariance with Generalized Means}
\label{app_geom_agg}
The two-level message passing scheme on any node $v_i$ in a hypergraph $\mathcal{H}=(\mathcal{V}, \mathcal{E}, \mathbf{X})$ comprises intra-edge aggregation function (within an edge $\mathbf{e}$) $\mathcal{F}_1^{(\mathbf{e})}(\cdot)$ and inter-edge aggregation function $\mathcal{F}_2(\cdot)$. For the choice of  generalized means, these can be stated as 
\begin{align*}
\mathcal{F}_1^{(\mathbf{e})}(\mathbf{s}_1) &= \left(  \frac{1}{|\mathcal{N}(v_i)|} \sum_{v_j \in \mathcal{N}(v_i,\mathbf{e})}  w_j \mathbf{x}_{j}^{p_1}\right)^{\frac{1}{p_1}} & \mathcal{F}_2(\mathbf{s}_2) &= \left(  \frac{1}{|E(v_i)|} \sum_{\mathbf{e}\in E(v_i)}  {\mathbf{s}_2}^{p_2}\right)^{\frac{1}{p_2}}.
\end{align*}
To reiterate here, $\mathbf{s}_1$  denotes the unordered set of input $\{\mathbf{x}_{j}\: | \: v_j \in \mathcal{N}(v_i,\mathbf{e}; \alpha)\}$ and $\mathbf{s}_2$ denotes the unordered set $\{ \mathcal{F}_1^{(\mathbf{e})}\,(\mathbf{s}_1) \enskip | \enskip \mathbf{e} \in E(v_i)\}$ as given in Equations 1-3 of the main paper. Here and henceforth in this proof, we remove the superscript index ‘$(\mathbf{e})$’ for the sake of clarity.

\emph{Theorem. } For  generalized means used as intra-edge and inter-edge aggregation functions, global neighborhood invariance is satisfied at any node $v_i$ if, $p1 = p2$ and $w_j = \frac{1}{|\mathcal{N}(v_i,\mathbf{e}_q)|} * \left(\sum_{m=1}^r{\frac{1}{|\mathcal{N}(v_i,\mathbf{e}_m)|}}\right)^{-1} \enskip \forall \enskip \mathbf{e} \in E(v_i, \mathbf{e}_q)$.



\emph{Proof. } The nested aggregation function can be stated as

\begin{equation}
        \mathcal{F}_2(\mathbf{s}_2) = \left(  \frac{1}{|E(v_i)|} \sum_{\mathbf{e}\in E(v_i)} 
       \left(   \frac{1}{|\mathcal{N}(v_i,\mathbf{e})|} \sum_{v_j \in \mathcal{N}(v_i,\mathbf{e})}  \mathbf{x}_{j}^{p_1}\right)^{\frac{p_2}{p_1}}    \right)^{\frac{1}{p_2}}
       \label{inter_main}
\end{equation}

 Equation \ref{inter_main} can be rewritten as 
\small
\begin{equation}\label{expanded}
        \mathcal{F}_2(\mathbf{s}_2) = \left(  \frac{1}{|E(v_i)|} \left(
        \left(   \frac{1}{|\mathcal{N}(v_i,\mathbf{e}_q)|} \sum_{v_j \in \mathcal{N}(v_i,\mathbf{e}_q)}  \mathbf{x}_{j}^{p_1}\right)^{\frac{p_2}{p_1}} + 
        \sum_{\mathbf{e}\in E(v_i),\mathbf{e}\neq \mathbf{e}_q} 
       \left(   \frac{1}{|\mathcal{N}(v_i,\mathbf{e})|} \sum_{v_j \in \mathcal{N}(v_i,\mathbf{e})}  \mathbf{x}_{j}^{p_1}\right)^{\frac{p_2}{p_1}}\right)    \right)^{\frac{1}{p_2}}
\end{equation}
\normalsize
Further, let 

\begin{equation}
    \Psi = \sum_{\mathbf{e}\in E(v_i),\mathbf{e}\neq \mathbf{e}_q} 
       \left(   \frac{1}{|\mathcal{N}(v_i,\mathbf{e})|} \sum_{v_j \in \mathcal{N}(v_i,\mathbf{e})}  \mathbf{x}_{j}^{p_1}\right)^{\frac{p_2}{p_1}},
\end{equation}
then Eq. \ref{expanded} can be rewritten as

\begin{equation}
        \mathcal{F}_2(\mathbf{s}_2) = \left(  \frac{1}{|E(v_i)|} \left(
        \left(   \frac{1}{|\mathcal{N}(v_i,\mathbf{e}_q)|} \sum_{v_j \in \mathcal{N}(v_i,\mathbf{e}_q)}  \mathbf{x}_{j}^{p_1}\right)^{\frac{p_2}{p_1}} + 
        \Psi\right)    \right)^{\frac{1}{p_2}}
\end{equation}

\begin{figure}
\centering
\begin{subfigure}{0.4\textwidth}
    \centering
     \includegraphics[scale=0.65]{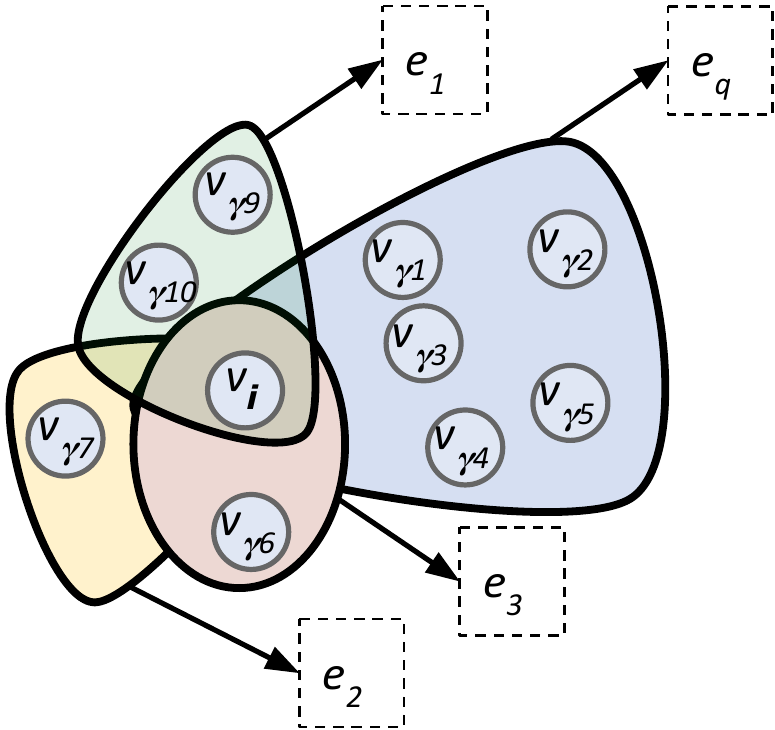}
    \caption{}
    \label{fig_expansion}
\end{subfigure}
\begin{subfigure}{0.4\textwidth}
    \centering
    \includegraphics[scale=0.55]{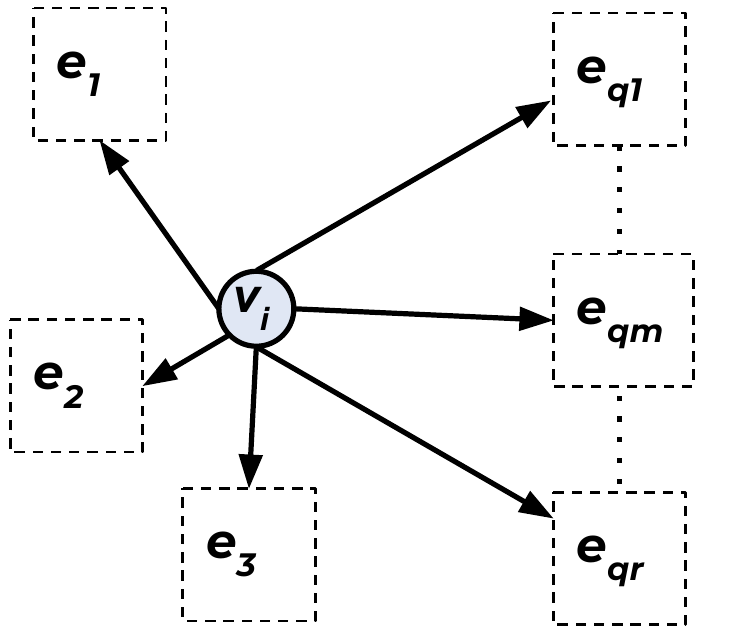}
    \caption{}
    \label{fig_split}
\end{subfigure}
\caption{(a) Example showing node $v_i$ shared across 4 hyperedges. (b) For reduced cardinality, hyperedge $e_q$ is split into $r$ hyperedges. Note that the global neighborhood of $v_i$ still remains the same, however its intra-edge neighborhood has changed due to such splitting. 
}
\label{appendix_split}
\end{figure}

Let us assume now that hyperedge $\mathbf{e}_q$ is split into $r$ hyperedges such that $v_i$ is the common node in all these hyperedges, given by  $E(v_i,\mathbf{e}_q)=\{\mathbf{e}_{q_1}, \mathbf{e}_{q_2} \hdots \mathbf{e}_{q_r}$\} as shown in Fig. \ref{appendix_split}. Stating the aggregation on the new set of hyperedges as $\Tilde{\mathcal{F}}_2(\mathbf{s}_2)$, we assemble the contribution from this new set of hyperedges with added weight terms $w_j$ as stated below.

\begin{equation}
        \Tilde{\mathcal{F}}_2(\mathbf{s}_2) = \left(  \frac{1}{|E(v_i)|} \left(
       \sum_{\mathbf{e}\in E(v_i,\mathbf{e}_q)} 
       \left(   \frac{1}{|\mathcal{N}(v_i,\mathbf{e})|} \sum_{v_j \in \mathcal{N}(v_i,\mathbf{e})}  w_j \mathbf{x}_{j}^{p_1}\right)^{\frac{p_2}{p_1}} + 
        \Psi\right)    \right)^{\frac{1}{p_2}}
\end{equation}

For the property of global neighborhood invariance to hold at $v_i$, the following condition should be satisfied:  $\mathcal{F}_2(\mathbf{s}_2) = \Tilde{\mathcal{F}}_2(\mathbf{s}_2)$. Based on this, we would like to solve for the weights $w_j$. For this, we equate the two terms and obtain
\begin{equation}
  \left(   \frac{1}{|\mathcal{N}(v_i,\mathbf{e}_q)|} \sum_{v_j \in \mathcal{N}(v_i,\mathbf{e}_q)}  \mathbf{x}_{j}^{p_1}\right)^{\frac{p_2}{p_1}} = \sum_{\mathbf{e}\in E(v_i,\mathbf{e}_q)} 
       \left(   \frac{1}{|\mathcal{N}(v_i,\mathbf{e})|} \sum_{v_j \in \mathcal{N}(v_i,\mathbf{e})}  w_j\mathbf{x}_{j}^{p_1}\right)^{\frac{p_2}{p_1}}
       \label{eq_global_eq}
\end{equation}

We further solve for the variables $p_1$, $p_2$ and $w_j$ where Eq. \ref{eq_global_eq} holds. For the sake of clarity, we first simplify Eq. \ref{eq_global_eq} using the following substitutions: $\alpha = \frac{p_2}{p_1}$, $\beta = \frac{1}{|\mathcal{N}(v_i,\mathbf{e}_q)|}$ and $\beta_{mj} = \frac{w_j}{|\mathcal{N}(v_i,\mathbf{e}_m)|}$, where the index $m$ here is used to refer to the $m^{\text{th}}$ hyperedge from among the $r$ hyperedges obtained on splitting $\mathbf{e}_q$. Further, let $z_j = \mathbf{x}_j^{p_1}$ for $v_j \in \mathcal{N}(v_i,\mathbf{e}_q)$ and $z_{mj} = \mathbf{x}_j^{p_1} $ for $v_j \in $$\mathcal{N}(v_i,\mathbf{e}_m)$ and $\mathbf{e}_m \in E(v_i,\mathbf{e}_q)$.

Based on these substitutions, Eq. \ref{eq_global_eq} can be restated as
\begin{align}
    \beta^{\alpha}(z_1 + z_2 + \hdots +z_N)^{\alpha} & = (\beta_{11}z_{1} + \beta_{12}z_{2} + \hdots + \beta_{1j}z_{j} + \hdots +\beta_{1N}z_{N})^{\alpha } \nonumber \\ 
    & + (\beta_{21}z_{1} + \beta_{22}z_{2} + \hdots + \beta_{2j}z_{j} + \hdots + \beta_{2N}z_{N})^{\alpha} + \nonumber \\
    &   \vdots   \nonumber\\
    & + (\beta_{r1}z_{1} + \beta_{r2}z_{2} + \hdots + \beta_{rj}z_{j} + \hdots + \beta_{rN}z_{N})^{\alpha}. 
    \label{eq_subs_eq}
\end{align}

We seek general solutions for $w_j$ and $\alpha$ which holds for all values of $z_j \in [0,1]$ since every element in the normalized feature vectors $\mathbf{x}_j$ lies in $[0,1]$. 

For a generalized solution, the coefficients of $z_{j}$ on the right should be equal to the coefficient of $z_j$ on the left. The term on the left can be reformulated as
\begin{equation} \label{lhs}
   \beta^{\alpha}(z_1 + z_2 + \hdots +z_N)^{\alpha}  = \beta^{\alpha}(z_1 + (z_2 + z_3+ \hdots +z_N))^{\alpha}  
\end{equation}

Consider the case when $|z_1| \leq |z_2+z_3+ \hdots|$, we expand Eq. \ref{lhs}. using binomial expansion for real co-efficients,
\begin{align}
    \beta^{\alpha}(z_1 + (z_2 + z_3+ \hdots ))^{\alpha} 
    &= \beta^\alpha({\binom{\alpha}{0}}z_1^\alpha +{ \binom{\alpha}{1}}z_1^{\alpha-1}(z_2 + z_3+ \hdots + z_N ) + 
    \nonumber \\ & \vdots 
    \nonumber \\ 
    &+ { \binom{\alpha}{\alpha -1 }}z_1(z_2 + z_3+ \hdots +z_N))
    \nonumber\\ 
    &= \beta^\alpha(z_1^\alpha + \alpha (z_1^{\alpha-1}z_2 + z_1^{\alpha-1}z_3+ \hdots + z_1^{\alpha-1}z_N) + 
    \nonumber \\ & \vdots  
    \nonumber\\   &+  \alpha z_1(z_2 + z_3+ \hdots + z_N )^{\alpha -1})  
    \label{eqn_bin_exp}
\end{align}

 Without any loss of generality, we consider splitting of hyperedge $e_q$ into $r$ hyperedges such that nodes 
 $v_{\gamma_1}$ and $v_{\gamma_2}$ shown in Fig. \ref{fig_expansion} are not contained in the same hyperedge anymore. This implies that RHS in \mbox{Eq. \ref{eqn_bin_exp}} should not contain product terms of $z_1$ and $z_2$. Hence,
 the term $z_1^{\alpha-1}z_2$ should be such that
 \begin{equation}
      \alpha - 1 = 0 \Rightarrow \alpha = 1 \Rightarrow p1 = p2
 \end{equation}

Substituting $\alpha =1$ and comparing the coefficients in Eq. \ref{eq_subs_eq}, we get
 \begin{align}
 \beta &= \beta_{11} + \beta_{12} + \hdots + \beta_{21} + \beta_{22} \hdots + \beta_{r1} + \beta_{r2} + \hdots  \nonumber \\
\frac{1}{|\mathcal{N}(v_i,\mathbf{e}_q)|} &=  \sum_{m=1}^r{\frac{w_j}{|\mathcal{N}(v_i,\mathbf{e}_m)|}} \\
w_j &= \frac{1}{|\mathcal{N}(v_i,\mathbf{e}_q)|} * \left(\sum_{m=1}^r{\frac{1}{|\mathcal{N}(v_i,\mathbf{e}_m)|}}\right)^{-1}
\end{align}

Thus, for $\mathbf{e} \in E(v_i, \mathbf{e}_q)$, if an edge $\mathbf{e}_q$ is split into multiple edges $E(v_i, \mathbf{e}_q)$, then for the two aggregations to hold, the conditions are $p_1 = p_2$ and $w_j = \frac{1}{|\mathcal{N}(v_i,\mathbf{e}_q)|} * \left(\sum_{m=1}^r{\frac{1}{|\mathcal{N}(v_i,\mathbf{e}_m)|}}\right)^{-1}  $.

\section{Analysis of connectedness in hypergraphs}
\begin{figure}%
\centering
\begin{tikzpicture}
    \centering
    \node[inner sep=0pt] (russell) at (0,0)
    {\includegraphics[scale=0.24]{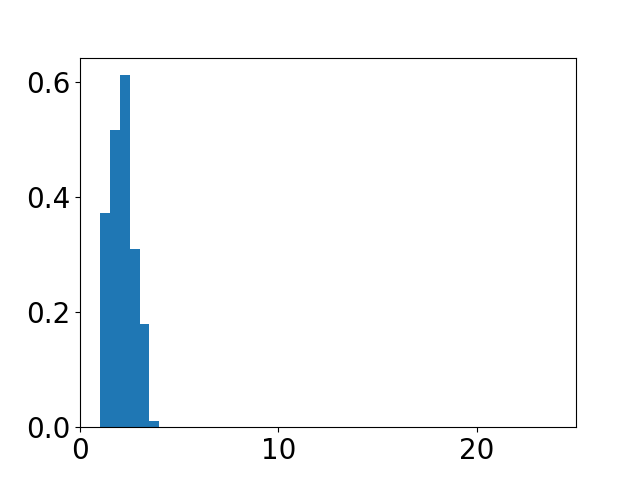}\hspace{-1em}\includegraphics[scale=0.24]{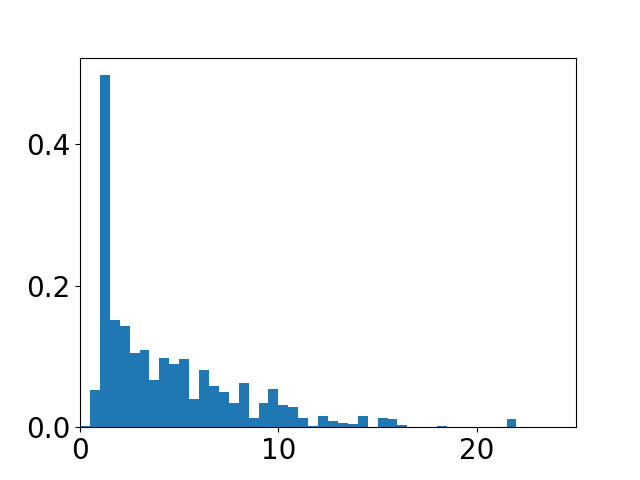}\hspace{-1em}\includegraphics[scale=0.24]{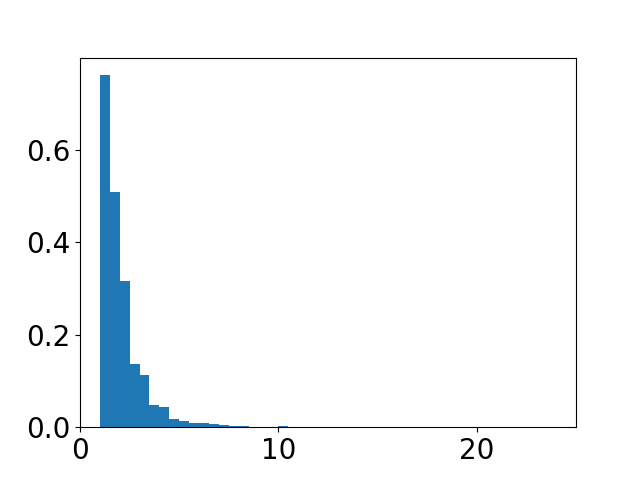}\hspace{-1em}\includegraphics[scale=0.24]{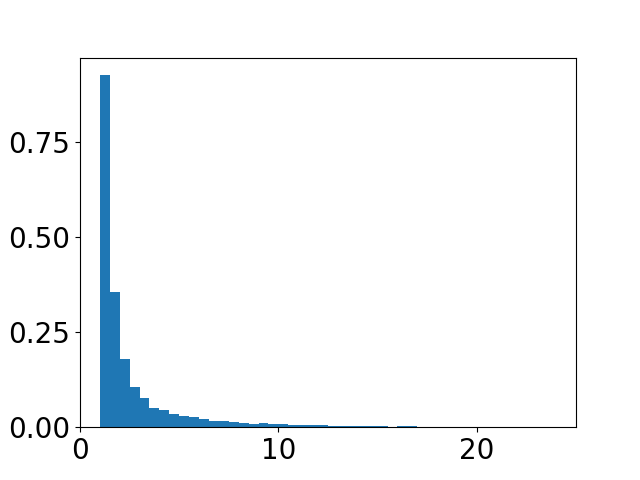}};
    \node[inner sep=0pt, rotate=90] (russell) at (-7.3,0){{\scriptsize Normalized frequency}};
    \node[inner sep=0pt] (russell) at (-5.3,-1.7){{\scriptsize $|\mathcal{N}(v)|/|E(v)|$}};
    \node[inner sep=0pt] (russell) at (-1.7,-1.7){{\scriptsize $|\mathcal{N}(v)|/|E(v)|$}};
    \node[inner sep=0pt] (russell) at (-1.7,1.4){{\scriptsize CORA co-authorship}};
    \node[inner sep=0pt] (russell) at (-5.3,1.4){{\scriptsize CORA co-citation}};
    \node[inner sep=0pt] (russell) at (1.9,1.4){{\scriptsize Citeseer}};
    \node[inner sep=0pt] (russell) at (5.3,1.4){{\scriptsize Pubmed}};
    \node[inner sep=0pt] (russell) at (1.9,-1.7){{\scriptsize $|\mathcal{N}(v)|/|E(v)|$}};
    \node[inner sep=0pt] (russell) at (5.3,-1.7){{\scriptsize $|\mathcal{N}(v)|/|E(v)|$}};
\end{tikzpicture}
     \caption{Distribution of $|\mathcal{N}(v)|/|E(v)|$ values across CORA co-citation, CORA-co-authorship , Citeseer and Pubmed datasets.}
    \label{data_analysis}%
\end{figure}
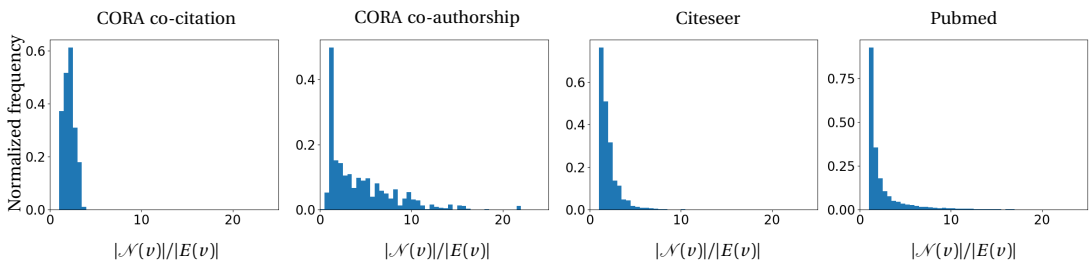

We analyze the importance of learning `connectedness' for hypergraphs as described in Section 3.3 of the main paper. We show example distributions of $|\mathcal{N}(v)|/|E(v)|$ for CORA co-citation, CORA co-authorship, Pubmed and Citeseer citation datasets. Fig. \ref{data_analysis} shows the histogram of $|\mathcal{N}(v)|/|E(v)|$ distributions on all these datasets . For the CORA co-citation dataset and Citeseer, $|\mathcal{N}(v)|/|E(v)|$ is mostly close to 1 or more compactly distributed while for the others (Cora co-authorship and Pubmed) the histogram shows a longer tail. We represent these datasets using both hypergraph as well as graph and apply the proposed HyperMSG (non-adaptive) model on them. Note that, for graph based data representation HyperMSG, simply converts to a message passing neural network on graphs and loses the two-level aggregation benefits. Table \ref{tab_graph} shows the performance comparison of HyperMSG on the two data structures. As can be seen,  the performance gain with message passing using HyperMSG over the graph-based model is relatively small for CORA co-citation dataset and Citeseer. On the contrary, for CORA co-authorship and Pubmed , where the distribution of $|\mathcal{N}(v)|/|E(v)|$ is right-skewed, the performance gain was significantly higher. This experiment illustrates that even with the same functional characteristics (features) of nodes, the structural information encoded within a hypergraph plays a key role in learning better node embeddings and provides a motivation for utilizing it by using a ''conenctedness" function.

\label{sec_app_en}

\begin{table}
\centering
\caption{Performance scores (in terms of accuracy \%) for HyperMSG (non-adaptive) on 4 datasets represented using graph as well as hypergraph. The performance shows the advantage of hypergrpahs over graphs for datasets with large variations in  $|\mathcal{N}(v)|/|E(v)|$ distribution.}
\begin{tabular}{lcc}
\toprule
  Dataset &Graph-based & Hypergraph-based \\
\cmidrule(l){1-1} \cmidrule(l){2-2} \cmidrule(l){3-3} 
Cora (co-citation) & 64.2 $\pm$ 2.3 & 69.3 $\pm$ 1.9  \\ [3pt]
Cora (co-authorship) & 73.9 $\pm$ 1.3 & 74.9 $\pm$ 1.0 \\[3pt]
Citeseer & 64.8 $\pm$ 0.8 & 66.4 $\pm$ 1.8  \\ [3pt]
Pubmed &  73.8 $\pm$ 1.8 & 76.2 $\pm$ 1.6\\
\bottomrule
\end{tabular}
\label{tab_graph}
\end{table}

\section{Experiments: Additional details}

 \subsection{Semi-supervised Node Classification}
 We perform multi-class classification on co-authorship and co-citation datasets, where the task is to predict the topic (class) for each document. Hypergraphs are created on these datasets by assigning each document as a node and each hyperedge represents (a) all documents co-authored by an author in co-authorship dataset and (b) all documents cited together by a document in co-citation dataset. Each document (node) is represented by bag-of-words features.
 \label{app_sec_dataset}
 
\textbf{Dataset Description}
 The details about nodes, hyperedges and features is shown in Table \ref{dataset}.  We use the same dataset and train-test splits as provided by \cite{yadati2019hypergcn} in their publically available implementation \footnote{HyperGCN Implementation: https://github.com/malllabiisc/HyperGCN}.
\begin{table}
\centering
\caption{Details of real-world hypergraph datasets used in our work}
\begin{tabular}{lccccc}
\toprule
  &DBLP &Pubmed &Citeseer & Cora &ArXiv \\
\cmidrule(l){2-2} \cmidrule(l){3-3} \cmidrule(l){4-4} \cmidrule(l){5-5} \cmidrule(l){6-6}
Nodes ($|\mathcal{V}|$) & 43413  & 19717 & 3312 & 2708 & 84893 \\ [3pt]
Hyperedges ($|\mathcal{E}|$) & 22535  & 7963 & 1079 & 1579 & 81994 \\[3pt]
average hyperedge size &4.7$\pm$6.1   & 4.3 $\pm$ 5.7 &3.2$\pm$2.0 & 3.0 $\pm$ 1.1 & 6.8  $\pm$ 5.2  \\ [3pt]
number of features, $|\mathbf{x}|$ &1425&500&3703&1433 & 1000 \\ [3pt]
number of classes &6&3&6&7& 6\\
Train:Test &0.044& 0.004& 0.043&0.054&0.050\\
\bottomrule
\end{tabular}
\label{dataset}
\end{table}

\textbf{Implementation details}
\label{app_sec_imp_details}
Experiments were run for 250 epochs on a GTX 1080 Ti with 12 GB RAM. The specific hyperparameter values used in this experiment can be found in Table \ref{imp_node}.

\begin{table}
\centering
\caption{Implementation details for Semi-supervised Node Classification}
\begin{tabular}{lc}
\toprule
  Hyperparameter &Value  \\
\cmidrule(l){1-1} \cmidrule(l){2-2} 
Hidden layers [size] & 2 [16]   \\ [3pt]
Dropout rate & 0.5  \\[3pt]
Learning rate & 0.01   \\ [3pt]
Weight decay & 0.0005\\
Epochs & 250\\
\bottomrule
\end{tabular}
\label{imp_node}
\end{table}

\subsection{Hypergraph classification on neuroimaging data}


\textbf{Dataset Description}
We used the Autism Brain Imaging Data Exchange Dataset (ABIDE) dataset for this experiment. The ABIDE (I+II) dataset is a collection of structural (T1w) and functional (rs-fMRI) brain images aggregated across 29 institutions  \cite{di2014autism}, available for download  \footnote{http://fcon\_1000.projects.nitrc.org/indi/abide/}. The dataset consists of brain images corresponding to 1,028 participants with a diagnosis of autism, Asperger or pervasive developmental disorder-not otherwise specified (called ASD), and 1,141 typically developing participants (CON). Some of these brain images, however fails to adhere to meet the suggested duration ( $\sim$ 4–5 min) for obtaining robust rs-fMRI estimates \cite{van2010intrinsic}.  We followed a preprocessing strategy adopted by the Preprocessed Connectome Project initiative \footnote{http://preprocessed-connectomes-project.org/abide/}. Further, certain common noise removal strategies such as motion correction and image normalization were performed using the same preprocessing pipeline as proposed by  \cite{thomas2020classifying}. Following the same quality check and subject selection as in \cite{thomas2020classifying}, we selected 1,162 subjects in total, out of which there were 620 ASD  and 542 CON subjects.


\textbf{Spatial Independent Components as Hypergraphs.}  The rsfMRI data from all subjects were used to perform an independent component analysis (ICA) at the group level to discover spatial patterns that were overlapping \cite{mckeown1998analysis}. The number of spatial components parcellate the brain into $N = 50$ regions. These $N$ regions each contain a few thousand voxels which are the represented by the nodes of the brain hypergraph. The brain parcellations overlap in space forming hyperedges. The total number of nodes encapsulated by these hyperedges were 22,879 and the time course (signal) belonging to each node was used as its feature vector. Thus, we get 1,162 brain hypergraphs ($\mathcal{H}$), each consisting 22,879 nodes ($\mathcal{V}$), 50 hyperedges ($\mathcal{E}$) and $22,879 \times 195$ dimensional feature matrix $(\mathbf{X})$.

\textbf{Implementation Details} We perform binary classification on the hypergraphs with a 5 fold cross validation. To learn representations for each hypergraph, we include fully connected layers at the end. The input to the fully connected layer is flattened out node representation vectors. 
Experiments were run for 250 epochs on a GTX 1080 Ti with 12 GB RAM. The specific hyperparameter values used in this experiment can be found in Table \ref{imp_brain}.
\begin{table}
\centering
\caption{Implementation details for Hypergraph clssification on neiroimaging data}
\begin{tabular}{lc}
\toprule
  Hyperparameter &Value  \\
\cmidrule(l){1-1} \cmidrule(l){2-2} 
Hidden layers [size] & 2 [64,16]   \\ [3pt]
Fully-connected layers [size] & 2 [704,256]   \\ [3pt]
Dropout rate & 0.5  \\[3pt]
Learning rate & 0.005   \\ [3pt]
Weight decay & 0.00005\\
Epochs & 100\\
\bottomrule
\end{tabular}
\label{imp_brain}
\end{table}

\section{Time Complexity Analysis}
Let $\mathcal{H}=(\mathcal{V}, \mathcal{E}, \mathbf{X})$ denote an attributed attributed hypergraph, where $\mathbf{X}$ is the feature matrix with each row representing node feature vectors. Let $d$ denote the dimension of the input feature vector, $h$ be the number of hidden layers and  $c$ the number of classes.  Further, let $T$ be the total number of training iterations and  $N = \sum_{\mathbf{e} \in \mathcal{E}} |\mathbf{e}|$; $N_m = \sum_{\mathbf{e} \in \mathcal{E}} 2|\mathbf{e}|-3 $; $N_l = \sum_{\mathbf{e} \in \mathcal{E}} \Comb{|\mathbf{e}|}{2} $, then
\begin{itemize}
    \item HyperMSG takes $\mathcal{O}(TN(1+h(d+c)))$ time
    \item HyperGCN \cite{yadati2019hypergcn} takes $\mathcal{O}(T(N+N_mh(d+c)))$
    \item HGNN \cite{feng2019hypergraph} takes $\mathcal{O}(TN_lh(d+c))$
\end{itemize}

Thus, in terms of complexity, HyperMSG is at par with the current state-of-the-art models HyperGCN and HGNN.

\section{Stability test on Cora and Citeseer}
\label{app:result}
\pgfplotsset{every axis/.append style={
                    label style={font=\Large},
                    tick label style={font=\Large}  
                    }}
\begin{figure} [H]
\centering
	\begin{subfigure}{0.35\linewidth}
	\centering
\begin{tikzpicture}[scale = 0.6]
  \begin{axis}
  [%
           title = {{\Large {Cora}}},
        		thick,
            legend pos = south east,
            xlabel = {Train:Test Ratio},
            ylabel = {Accuracy ($\%$)},
            ymin=55, ymax=90,
            xtick={0,1,2,3,4},
    xticklabels={$1/30$,$1/20$,$1/15$,$1/10$,$1/3$}
            ]
  
    \addplot [solid, color = blue, mark=asterisk, error bars/.cd, y dir=both, y explicit,
      error bar style={line width=2pt,solid},
      error mark options={line width=1pt,mark size=4pt,rotate=90}]
    table [x=x, y=y, y error=y-err]{%
      x y y-err
      0 73.2 2.3
      1 77.7 1.2
      2 79.5 1.1
      3 81.1 1.0
      4 81.5 0.7
    };
    \addplot [solid,color = red, mark=asterisk, error bars/.cd, y dir=both, y explicit,
      error bar style={line width=2pt,solid, color = red},
      error mark options={line width=1pt,mark size=4pt,rotate=90}]
    table [x=x, y=y, y error=y-err]{%
      x y y-err
      0 66.1 4.1
      1 69.7 3.0
      2 71.3 2.3
      3 73.6 1.9
      4 74.8 1.3
    };
    \addlegendentry{HyperMSG}
  \addlegendentry{HyperGCN}

  \end{axis}
 \end{tikzpicture}
 \end{subfigure}
 \hspace{-0.2cm}
\begin{subfigure}{0.35\linewidth}
\centering
\begin{tikzpicture}[scale = 0.6]
  \begin{axis}
  [%
           title = {{\Large {Citeseer}}},
        		thick,
            legend pos = south east,
            xlabel = {Train:Test Ratio},
            ylabel = {Accuracy ($\%$)},
            ymin=45, ymax=85,
    xtick={0,1,2,3,4},
    xticklabels={$1/50$,$1/30$,$1/25$,$1/10$, $1/3$},
            ]
    \addplot [solid, color = blue,mark=asterisk, error bars/.cd, y dir=both, y explicit,
      error bar style={line width=2pt,solid},
      error mark options={line width=1pt,mark size=4pt,rotate=90}]
    table [x=x, y=y, y error=y-err]{%
      x y y-err
      0 63.2 2.5
      1 64.6 2.2
      2 66.8 1.6
      3 68.3 1.3
      4 68.9 1.3
    };
    \addplot [solid,color = red, mark=asterisk, error bars/.cd, y dir=both, y explicit,
      error bar style={line width=2pt,solid, color = red},
      error mark options={line width=1pt,mark size=4pt,rotate=90}]
    table [x=x, y=y, y error=y-err]{%
      x y y-err
      0 60.5 5.5
      1 61.8 5.4
      2 62.7 4.6
      3 63.9 4.5
      4 65.2 4.0
    };
            \addlegendentry{HyperMSG}
       \addlegendentry{HyperGCN}
  \end{axis}
 \end{tikzpicture}
 \end{subfigure}
     \caption{Accuracy scores for HyperMSG and HyperGCN obtained for different train-test ratios. }
    \label{fig_stable_cora}%
 \end{figure}
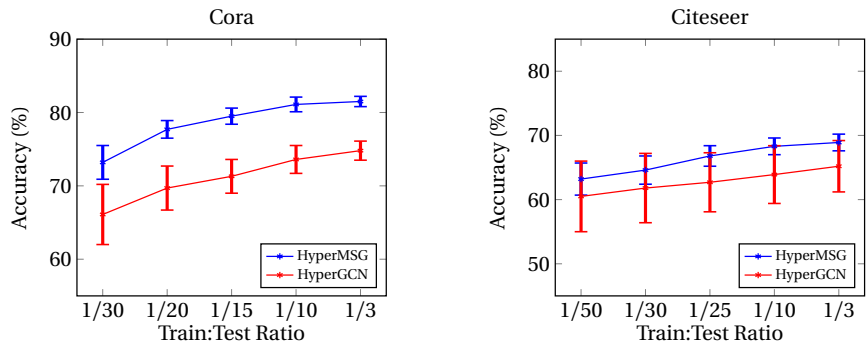

In the main text, we showed the stability of HyperMSG on Pubmed and DBLP datasets through analyzing the performance on different train-test ratios. Here, we show similar results for Cora and Citeseer datasets as well.
We  study the stability of our method in terms of the variance observed in performance for different train-test split ratios. 
Fig. \ref{fig_stable_cora} shows results for the two learning methods (HyperGCN and HyperMSG) on 5 different train-test ratios. We see that the performance of both models improves when a higher fraction of data is used for training, and the performances are approximately the same at the train-test ratio of 1/3. However, for smaller ratios, we see that HyperMSG outperforms HyperGCN by a significant margin across all datasets. Further, the standard deviation for the predictions of HyperMSG 
is significantly lower than that of HyperGCN. Clearly, this implies that HyperMSG is able to better exploit the information contained in the hypergraph compared to HyperGCN, and can thus produce more accurate and stable predictions. 

\end{document}